\documentclass[lettersize,journal]{IEEEtran}
\usepackage[switch]{lineno}
\usepackage{amsmath,amsfonts}
\usepackage{algorithmic}
\usepackage{algorithm}
\usepackage{array}
\usepackage[caption=false,font=normalsize,labelfont=sf,textfont=sf]{subfig}
\usepackage{textcomp}
\usepackage{stfloats}
\usepackage{url}
\usepackage{verbatim}
\usepackage{graphicx}
\usepackage{cite}
\usepackage{multirow}
\usepackage{booktabs}
\usepackage{amsbsy}
\usepackage{rotating}
\usepackage{bigstrut}
\usepackage{color}
\usepackage[table]{xcolor}
\usepackage{tikz,orcidlink}
\hypersetup{
	colorlinks=true,
	linkcolor=black,
	citecolor=black
}
\let\oldequation\equation
\let\oldendequation\endequation
\renewenvironment{equation}{\linenomathNonumbers\oldequation}{\oldendequation\endlinenomath}

\begin{document}

\title{Unified-modal Salient Object Detection via Adaptive Prompt Learning}

\author{Kunpeng Wang\hspace{-1.5mm}$^{~\orcidlink{0000-0002-2788-7583}}$, Chenglong Li\hspace{-1.5mm}$^{~\orcidlink{0000-0002-7233-2739}}$, Zhengzheng Tu\hspace{-1.5mm}$^{~\orcidlink{0000-0002-9689-8657}}$, Zhengyi Liu\hspace{-1.5mm}$^{~\orcidlink{0000-0003-3265-823X}}$, and Bin Luo\hspace{-1.5mm}$^{~\orcidlink{0000-0001-5948-5055}}$,~\IEEEmembership{Senior~Member,~IEEE} 
	\thanks{This work was supported in part by the National Natural Science Foundation of China under Grant 62376005, 61876002, in part by the Natural Science Foundation of Anhui Higher Education Institution of China under Grant KJ2020A0033, in part by Anhui Provincial Natural Science foundation under Grant 2108085MF211, in part by Anhui Energy Internet Joint Fund Project under Grant 2008085UD07, in part by Anhui Provincial Key Research and Development Program under Grant 202104d07020008. (Corresponding author is Zhengzheng Tu and Bin Luo)}
	\thanks{Kunpeng Wang, Zhengzheng Tu and Bin Luo are with 
		Information Materials
		and Intelligent Sensing Laboratory of Anhui Province, Anhui
		Provincial Key Laboratory of Multimodal Cognitive Computation,
		School of Computer Science and Technology, Anhui University, Hefei
		230601, China (e-mail: kp.wang@foxmail.com; zhengzhengahu@163.com and luobin@ahu.edu.cn)}
	\thanks{Zhengyi Liu is with Key Laboratory of Intelligent Computing and Signal Processing of Ministry of Education, School of Computer Science and Technology, Anhui University, Hefei, China (e-mail: liuzywen@ahu.edu.cn)}
	\thanks{Chenglong Li is with Anhui Provincial Key Laboratory of Multimodal
	Cognitive Computation, School of Artificial Intelligence, Anhui University,
	Hefei 230601, China, and also with the Institute of Physical Science and
	Information Technology, Anhui University, Hefei 230601, China (e-mail:
	lcl1314@foxmail.com)}}

\markboth{Journal of \LaTeX\ Class Files,~Vol.~14, No.~8, August~2021}%
{Shell \MakeLowercase{\textit{et al.}}: A Sample Article Using IEEEtran.cls for IEEE Journals}


\maketitle

\begin{abstract}
Existing single-modal and multi-modal salient object detection (SOD) methods focus on designing specific architectures tailored for their respective tasks. However, developing completely different models for different tasks leads to labor and time consumption, as well as high computational and practical deployment costs. In this paper, we attempt to address both single-modal and multi-modal SOD in a unified framework called UniSOD, which fully exploits the overlapping prior knowledge between different tasks. Nevertheless, assigning appropriate strategies to modality variable inputs is challenging. To this end, UniSOD learns modality-aware prompts with task-specific hints through adaptive prompt learning, which are plugged into the proposed pre-trained baseline SOD model to handle corresponding tasks, while only requiring few learnable parameters compared to training the entire model. Each modality-aware prompt is generated from a switchable prompt generation block, which adaptively performs structural switching based on single-modal and multi-modal inputs without human intervention. Through end-to-end joint training, UniSOD achieves overall performance improvement on 14 benchmark datasets for RGB, RGB-D, and RGB-T SOD, which demonstrates that our method effectively and efficiently unifies single-modal and multi-modal SOD tasks. The code and results are available at \href{https://github.com/Angknpng/UniSOD}{https://github.com/Angknpng/UniSOD}.
\end{abstract}


\section{Introduction}
\begin{figure}[t]
	\centering
	\includegraphics[width=1\columnwidth]{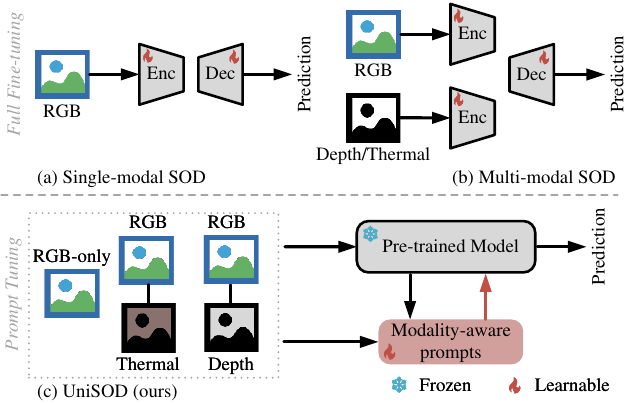}
	\caption{Workflow comparisons between existing Salient object detection (SOD) models and our UniSOD. \textbf{(a) \& (b)}: Existing single-modal and multi-modal SOD methods fully fine-tune the entire encoder-decoder based model, which is designed with a specific architecture tailored for the corresponding SOD task.  \textbf{(c)}: The proposed UniSOD handles both single-modal and multi-modal inputs in a unified framework, which learns a few parameters of modality-aware prompts to drive a frozen pre-trained model to address corresponding SOD tasks.} 
	\label{fig::motivation}
\end{figure}

Salient object detection (SOD) aims to identify and segment object(s) that attract human attention in a visible image, and its workflow is shown in Fig.~\ref{fig::motivation} (a). It can help eliminate redundant information and has been applied in many computer vision tasks, such as instance segmentation~\cite{li2023boost}, object tracking~\cite{zhang2020track}, and video analysis~\cite{cong2019video,fan2019video}. Though achieving great success~\cite{zhou2023texture,ma2023bbrf,tian2023modeling}, single-modal SOD still fails in some challenging scenes, e.g., low illumination, complex background, and similar foreground and background. 
To tackle these challenges, depth maps and thermal images have been successively introduced for the single-modal SOD to provide additional geometric spatial information and overall shape information in various complex scenes, respectively. Therefore, multi-modal SOD mainly exploits the complementary benefits of RGB-Depth (RGB-D)~\cite{piao2019depth,Zhao19CPFP,fu2020JLDCF,zhang20UCNet,liu20Learning} or RGB-Thermal (RGB-T)~\cite{tu2019rgb,tu2020rgbt,tu2021multi,liu2022swinnet} image pairs for robust prediction, as shown in Fig.~\ref{fig::motivation} (b). 

\begin{figure}[t]
	\centering
	\includegraphics[width=1\columnwidth]{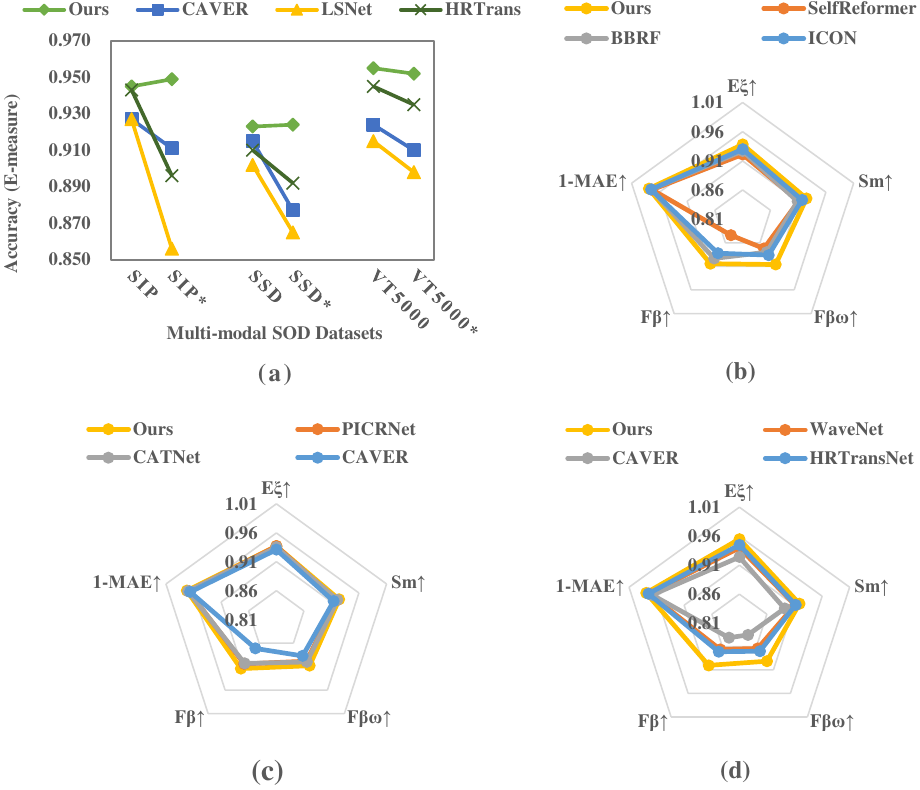}
	\caption{\textbf{(a)}: Performance variation of our method and advanced multi-modal methods~\cite{pang2023caver,zhou2023lsnet,Tang2023RTransNet} with modal completeness and absence. X* represents the absence of depth or thermal modal for the X dataset. \textbf{(b) (c) (d)} show the performance comparison of our method with advanced RGB-only SOD methods~\cite{Yun23SelfReformer,ma2023bbrf,Mingchen23ICON}, RGB-D SOD methods~\cite{Cong23PICRNet,Sun23CATNet,pang2023caver}, and RGB-T SOD methods~\cite{zhou23WaveNet,pang2023caver,Tang2023RTransNet} on the DUT-TE dataset~\cite{wang2017Learning}, STERE dataset~\cite{niu2012leveraging}, and VT5000 dataset~\cite{tu2020rgbt}, respectively. Overall, our method is slightly affected by modal absence and achieves superior performance in both single-modal and multi-modal SOD tasks.} 
	\label{fig::performance}
\end{figure}

However, in real-world applications, depth and thermal infrared sensors are not widely available~\cite{zhao2022joint,Casser2019depth,xuan2021thermal,Anjith2020thermal}, which results in the absence of corresponding depth or thermal modalities. In this case, multi-modal models can only rely on the RGB modality and suffer from performance degradation, as shown in of Fig.~\ref{fig::performance} (a).
To preserve performance, a straightforward notion is to train an extra single-modal SOD model from scratch only for visible images. In this way, different models need to be designed and trained separately for single-modal and multi-modal inputs. For example, Liu et al.~\cite{liu2021vst} design a framework for both RGB and RGB-D SOD based on vision transformer. Different from the RGB SOD model, they use an exclusive cross-modal transformer for the RGB-D SOD model to perform multi-modal information interaction. Zhou et al.~\cite{zhou2023texture} form an unsupervised framework to aggregate appearance information in single-modal or multi-modal SOD tasks through corresponding boundary-aware matching and saliency distilling strategies. 

Although effective, in order to handle single-modal and multi-modal inputs, these methods require the design of different specific networks to train the corresponding models, which is laborious and increases the computational and deployment burden, especially for modern transformer-based methods~\cite{Mingchen23ICON,liu2022swinnet,Cong23PICRNet} with more parameters. In addition, with limited training data for an individual task, task-specific models are prone to overfitting specific data distributions, resulting in sub-optimal performance, as shown in (b) (c) (d) of Fig.~\ref{fig::performance}. Therefore, it is necessary to exploit the strengths of the multiple training data from single-modal and multi-modal SOD tasks to jointly train a unified model that can optimally address both single-modal and multi-modal SOD tasks. However, jointly training such a unified SOD model end-to-end is non-trivial. On the one hand, the model has to take advantage of the commonality of single-modal and multi-modal SOD. On the other hand, the model has to automatically assign different processing strategies to single-modal and multi-modal inputs to address the corresponding tasks.

Recently, prompt learning has attracted great attention as it boosts the performance of many Natural Language Processing (NLP) downstream tasks~\cite{Francine21Power,Liu23NLP} by only fine-tuning prompts. Some computer vision research~\cite{jia2022vpt,lee2023Multimodal,zhu2023vipt} also show the great potential of prompt learning to be an alternative to full fine-tuning, which requires tuning a large number of parameters by training a model from scratch. The key of prompt learning is to exploit the common representation capabilities of a frozen pre-trained model and learn task-specific cues to adapt the pre-trained model to downstream tasks. Inherently, single-modal and multi-modal SOD have different inputs but the same objective, implying the substantial overlap in prior knowledge regarding feature extraction and saliency cue mining. In addition, both single-modal and multi-modal SOD have to deal with the RGB modality, and these modalities (i.e., RGB, depth, and thermal modalities) share some commonalities~\cite{fu2020JLDCF}, such as usually displaying object regions that are prominent to the surroundings. Therefore, it is natural to introduce prompt learning on top of a pre-trained RGB-based SOD model to efficiently exploit the common prior knowledge of single-modal and multi-modal SOD.

To this end, we propose a unified SOD framework, called UniSOD, for both single-modal and multi-modal SOD through prompt learning. As shown in Fig.~\ref{fig::motivation} (c), for different input cases (i.e., RGB-only, RGB-D, RGB-T), UniSOD adaptively learns modality-aware prompts in an end-to-end manner to drive the pre-trained model to address the corresponding tasks uniformly. Specifically, we propose a simple but effective baseline SOD model that learns sufficient prior knowledge from the training set of single-modal SOD for pre-training.
In addition, we propose a switchable prompt generation (SPG) block, which adaptively performs structure switching without manual intervention to generate modality-aware prompts. Each modality-aware prompt is integrated with the intermediate features of the pre-trained baseline SOD model, with all parameters frozen, to facilitate multi-stage interaction. Through end-to-end joint training on single-modal and multi-modal SOD datasets, UniSOD learns few parameters of the prompts to adapt the pre-trained model to different SOD tasks, including RGB, RGB-D, and RGB-T SOD. The main contributions of our work are summarized as follows:
\begin{itemize}
	\item We propose a novel unified framework for RGB, RGB-D, and RGB-T SOD through prompt learning, which adapts the pre-trained model to both single-modal and multi-modal SOD tasks with few learnable parameters.
	\item We design a simple and effective baseline SOD model for pre-training, which provides rich prior knowledge for both single-modal and multi-modal SOD tasks.
	\item We design a switchable prompt generation block to generate modality-aware prompts, which adaptively performs structure switching based on single-modal and multi-modal inputs without manual intervention.
	\item Extensive experiments on 14 benchmark datasets demonstrate the superior performance of our proposed UniSOD, with maximum improvement of 29.4\%, 14.3\%, and 19.0\% on $MAE$ metric for RGB, RGB-D, and RGB-T SOD, respectively, proving the effectiveness and potential of the proposed unified framework.
\end{itemize}

\section{Related Work}
\subsection{Salient Object Detection}
General salient object detection (SOD) aims to recognize and segment the most attractive objects from visible images. With the development of deep neural networks~\cite{simonyan2015VGG,he2016resnet,gao2021res2net}, numerous studies~\cite{wang2022survey} have been presented based on boundary enhancement~\cite{wang2019salient,zhao2019egnet,qin2019basnet,yun2022Recursive}, feature refinement~\cite{Wu19CPD,Zhao20GateNet,Wu22EDN,liu2023poolnet}, multi-decoder~\cite{Wei20LDF,Wei20F3Net}, attention mechanism~\cite{liu2020picanet,zhou2020multiattention,Wang23MENet}, and uncertainty perception~\cite{li2021uncertainaware,wang2022multi,tian2023modeling}. Recently, some methods~\cite{liu2021vst,xie2022Pyramid,ma2023bbrf} with impressive performance have been developed due to the long-range dependency modeling capabilities of Transformer~\cite{vaswani2017Attention}. However, these methods still struggle to tackle some challenging scenes, such as complex and low-contrast backgrounds, illumination changes, blur, etc. To deal with these inherent challenges, some studies introduce depth~\cite{piao2019DepthInduced,fu2020JLDCF,Ji21DCF,sun2021dsa2f,zhang20UCNet,ji2022promoting,pang2023caver} or thermal~\cite{tu2020rgbt,tu2021multi,liu2022swinnet,tu2022weakly,zhou2023lsnet,Tang2023RTransNet} images as complementary information for visible images. For example, Piao et al.~\cite{piao2019DepthInduced} integrate multi-scale complementary cues from RGB and depth streams for accurate salient object location in complex background scenes. Sun et al.~\cite{sun2021dsa2f} utilize geometric priors in the depth map to enhance RGB features and reduce background interference, thus accurately capturing salient objects in complex scenes. To preserve performance in challenging scenes, Tu et al.~\cite{tu2021multi} fuse features of RGB and thermal modalities through a multi-interactive dual decoder.

Nonetheless, the limited availability of depth and thermal sensors in practice leads to degraded performance of these multi-modal models, which is even inferior to that of single-modal ones~\cite{lee2023Multimodal,ge2023MetaBEV,Konwer2023ICCV}. Therefore, some recent studies~\cite{liu2021vst,zhou2023texture} tend to design methods that address both single-modal and multi-modal SOD, but require different structural designs and duplicated training for different tasks. For example, Liu et al.~\cite{liu2021vst} design a transformer-based SOD framework, in which the RGB-D SOD model incorporates a cross-modal converter on top of the RGB SOD model for multi-modal information interaction. In addition, some methods~\cite{zhao2022joint,chen2020improved,Zhang2022deep} attempt to estimate depth information from visible images as a complement to the missing modality. For example, Zhao et al.~\cite{zhao2022joint} estimate and purify the depth map to reduce the interference information in it, which improves the performance to some extent. However, the performance of these methods depends on the depth estimation module, which cannot guarantee the quality of the estimated features. In this paper, from a novel perspective, we utilize the capability of a pre-trained model through prompt tuning to address single-modal and multi-modal SOD uniformly, with a few parameters and low computational cost.

\subsection{Prompt Learning in Computer Vision}
\begin{figure*}[t]
	\centering
	\includegraphics[width=1\linewidth]{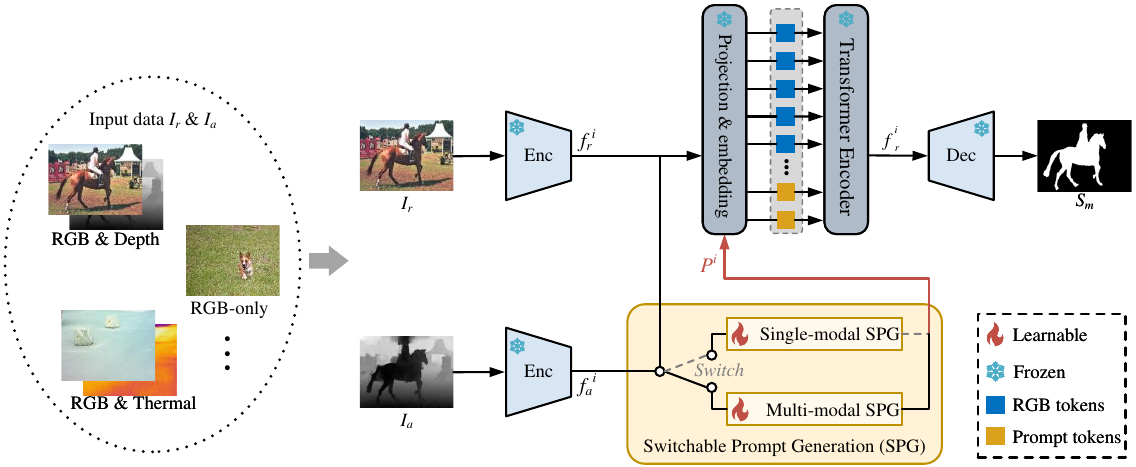}
	\caption{Overall architecture of our proposed UniSOD model for both single-modal and multi-modal SOD. The framework builds upon the proposed baseline SOD model (refer to Fig.~\ref{fig::RGB_framework}), where all parameters are pre-trained and frozen. Initially, the single-modal and multi-modal inputs are fed to the encoder for the extraction of multi-level features. Subsequently, a switchable prompt generation (SPG) block is designed to generate modality-aware prompts through adaptive structural switching without manual intervention. These prompts are attached to the intermediate features of the pre-trained model and then fed into the transformer encoder. This process facilitates prompt learning and drives the pre-trained model to address corresponding SOD tasks.} 
	\label{fig::framework}
\end{figure*}
In computer vision, full fine-tuning is widely employed to leverage the generalized representational capabilities of pre-trained models for downstream tasks. The common practice is to first train a common pre-trained model on a large-scale dataset, and then fine-tune the parameters of the entire pre-trained model in different downstream tasks. This approach results in satisfactory performance, but requires separate training and storage of all model parameters for different downstream tasks. 

Recently, prompt learning demonstrates its efficiency in the field of nature language processing, which appends a prompt to the input to instruct the pre-trained model for downstream tasks. Consequently, visual prompts are also introduced into many computer vision tasks~\cite{jia2022vpt,tu2023visual,dong2023LTP,zhu2023vipt,lee2023Multimodal} to adapt the pre-trained model to downstream vision tasks by tuning few learnable parameters. In order to implement prompt learning in vision models, VPT~\cite{jia2022vpt} introduces additional few learnable parameters in the input space of the frozen  pre-trained transformer model to adapt to downstream vision tasks. 
Unlike VPT that adapts to the pre-trained model, VQT~\cite{tu2023visual} introduces prompts to summarize frozen intermediate features, and the output features of the prompts are used for prediction. Pro-tuning~\cite{nie2023pro} introduces prompt blocks to generate visual cues, which are embedded into different stages of the pre-trained model to enrich the feature space.
For the application of prompt learning to downstream tasks, LPT~\cite{dong2023LTP} categorizes prompts into shared and specific ones to make the pre-trained model discriminative for long-tail data. ViPT~\cite{zhu2023vipt} introduces modality-complementary prompts to address different multi-modal tracking tasks. However, it only designs prompts for multi-modal inputs, which makes it difficult to handle the absence of auxiliary modalities. In addition, it needs to learn the parameters of modality-complementary cues for each multi-modal task, which increases the training and deployment burden. In this paper, we design a switchable prompt generation block to adaptively generate modality-aware prompts, which can simultaneously adapt the pre-trained model to both single-modal and multi-modal SOD tasks.

\section{Methodology}
In this paper, we propose UniSOD to address both single-modal and multi-modal SOD in a unified way. Instead of designing multiple specific models for different tasks and fully fine-tuning the entire model, UniSOD uniformly learns modality-aware prompts with few learnable parameters to drive the pre-trained model to address the corresponding tasks. 

\subsection{Overview}
\textbf{Problem Definition.} 
Given an visible image $I_{r}$ of size ${\mathbb{R}}^{H \times W}$, the objective of single-modal SOD is to predict a saliency map $S_{m}$, where each scalar value $S^{i,j}_m$ represents the confidence of a pixel belonging to the foreground region. For multi-modal SOD, it introduces an additional modality for the original RGB modality, extends the input to $\{ {I_r},{I_a}\}$, where $I_a$ represents depth or thermal image. The goal of multi-modal SOD is the same as that of single-modal SOD. Therefore, the key to unified SOD research is to make full use of their overlapping prior knowledge to efficiently address different SOD tasks.

\textbf{Overall Framework.}
Fig.~\ref{fig::framework} shows an overall architecture of the proposed UniSOD model. Given a pre-trained SOD model, which has strong representation and discrimination capabilities for various features, we keep all its parameters frozen to fully exploit the capabilities without impairing it. Considering the generalization ability of the pre-trained model, for input of any modality (i.e., $I_r$ and $I_a$), we all send it to the same encoder to extract multi-level features. After that, each level of features is fed into a switchable prompt generation block (SPG) to produce modality-aware prompts, which are then appended to the inherent input features. Therefore, the input space of the pre-trained transformer encoder is modified, and the parameters of the prompts are learned through the multi-layer transformer encoder and progressive decoder. In this way, for both single-modal and multi-modal inputs, the pre-trained model is driven by modality-aware prompts to predict the corresponding saliency map.

\subsection{Baseline SOD Model for Pre-training}
\begin{figure}[t]
	\centering
	\includegraphics[width=1\linewidth]{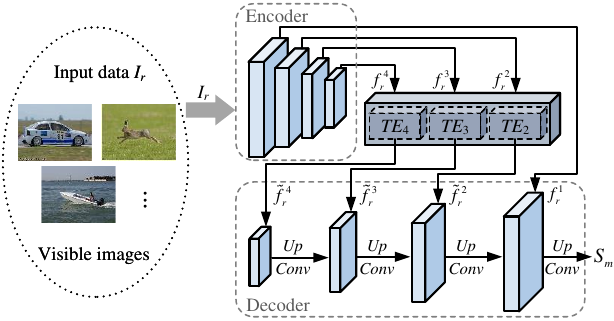}
	\caption{Detailed architecture of the pre-trained RGB SOD model.} 
	\label{fig::RGB_framework}
\end{figure}
The multi-modal SOD model can be viewed as an extension of the single-modal SOD (i.e., RGB SOD) model by adding a stream for the extra modality (i.e., depth or thermal modality) based on the original RGB stream, and the two streams use the same backbone in most methods~\cite{Zhou21SPNet,pang2023caver,liu2022swinnet}. In addition, both single-modal and multi-modal SOD have the same purpose of predicting the foreground region in the image(s) so as to obtain the final saliency map. Therefore, we attempt to use a pre-trained model with an RGB SOD framework to maximize the inclusion of prior knowledge common to both single-modal and multi-modal SOD. However, existing RGB SOD frameworks~\cite{qin2019basnet,pang20MINet,Wang23MENet} are almost designed for specific problems, which impairs the generalization ability of the pre-trained model to various kinds of data. Moreover, their framework structure is complex and is difficult to be extended to different downstream tasks. To this end, we design a simple and effective baseline SOD framework for pre-training, as shown in Fig.~\ref{fig::RGB_framework}.

\textbf{Model Architecture.}
The framework contains three components: an encoder for feature extraction, a transformer encoder for integrating the extracted features, and a decoder for feature reconstruction. Considering the strong feature representation ability of Swin Transformer~\cite{iccv2021swin}, we adopt it with pre-trained weights as the backbone in the encoder, the same as~\cite{liu2022swinnet}. In this way, the input image $I_r$ is first split into non-overlapping patches, and neighboring patches are merged as the backbone deepens. Then, hierarchical feature representations are constructed, denoted as $\left\{ {f_r^i} \right\}_{i = 1}^4$. 
To further integrate the extracted features, we flatten and feed them into a $L$-layer standard vision transformer encoder, which has the ability to model long-range dependencies. This process is formulated as: 
\begin{equation}\label{Eq::TE}
	\begin{split}
		&\tilde f_r^i = {\rm{reshape}}\left( {T{E_i}\left( {{\rm{flatten}}\left( {f_r^i} \right)} \right)} \right), \hfill i = 2,...,4,
	\end{split}
\end{equation} 
where $f_r^i$ is the output feature after reshaping. $T{E_i}$ denotes the $i_{th}$ level transformer encoder, which contains multi-head self-attention, layer norm, feed-forward network, and residual connection, and we refer readers to the literature~\cite{ICLR21transformer} for more detailed descriptions. Note that the extracted features of the first level are not fed into the transformer encoder due to the heavy computational burden of large resolution data. 
Subsequently, the decoder aggregates multi-level features in a top-down manner to progressively reconstruct high-resolution feature maps for final prediction, which can be formulated as:
\begin{equation}\label{Eq::decoder1}
	\renewcommand{\arraystretch}{1.0}
	\begin{split}
		&f_s^i = \left\{ \begin{gathered}
			Con{v_{3 \times 3}}(U{p_2}(\tilde f_r^i)), \hfill i = 4,\\
			Con{v_{3 \times 3}}(U{p_2}(\tilde f_r^i + f_s^{i + 1})), \hfill i = 2,3,\\
		\end{gathered} \right.
	\end{split}
	\renewcommand{\arraystretch}{1.0}
\end{equation}
\begin{equation}\label{Eq::decoder2}
	\begin{split}
		&{S_m} = Con{v_{3 \times 3}}\left(U{p_4}\left(Con{v_{3 \times 3}}\left(f_s^2 + f_r^1\right)\right)\right),
	\end{split}
\end{equation} 
where $S_m$ represents the predicted saliency map, $Con{v_{3 \times 3}}$ is a 3 × 3 convolutional layer, $U{p_x}$ is the $x$× upsampling operation with bilinear interpolation. We introduce the first-level features here to supplement detailed information.

\textbf{Optimization Objective.}
For the pre-trained model, we fine-tune all its parameters during the training process. Following~\cite{tu2021multi,Wu22MobileSal}, we use a combination of binary cross-entropy loss, smoothness loss~\cite{godard2017unsupervised}, and dice loss~\cite{milletari2016v} as the overall loss function to optimize our method,
\begin{equation}\label{Eq::loss}
	\begin{split}
		&{\cal L} = {{\cal L}_{bce}} + {{\cal L}_{smooth}} + {{\cal L}_{dice}},
	\end{split}
\end{equation} 
where ${\cal L}_{bce}$, ${\cal L}_{smooth}$, and ${\cal L}_{dice}$ denote the binary cross-entropy loss, smoothness loss, and dice loss, respectively, between the predicted saliency map $S_m$ and ground truth. 

\subsection{Switchable Prompt Generation}
\begin{figure}[t]
	\centering
	\includegraphics[width=1\linewidth]{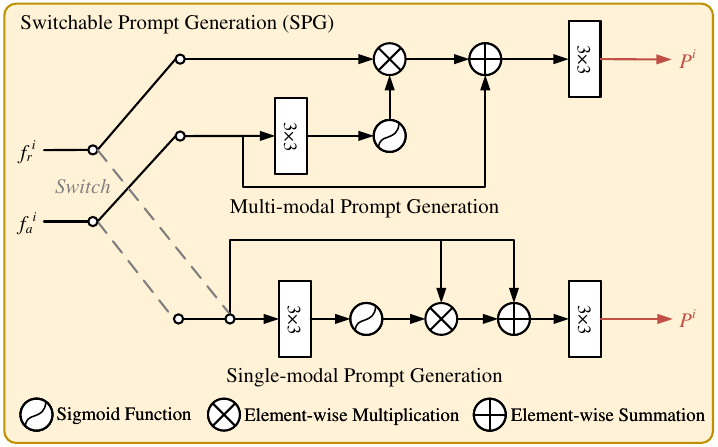}
	\caption{Detailed design of the proposed SPG block. The SPG adaptively switches the structure based on single-modal and multi-modal inputs to generate modality-aware prompts.} 
	\label{fig::SPG}
\end{figure}
Existing SOD methods almost exclusively design frameworks for single-modal or multi-modal SOD tasks, or design different frameworks for both. However, designing and training different models for different tasks is laborious and resource-intensive. Therefore, a model that uniformly and efficiently addresses both single-modal and multi-modal SOD is needed to cope with different situations in real-world applications. Some recent works~\cite{jia2022vpt,dong2023LTP,zhu2023vipt} introduce a small number of parameters to the frozen pre-trained model to learn visual prompts. By fine-tuning them during training, they achieves impressive results on many downstream tasks. However, these methods require multiple rounds of training to learn specific prompts for different tasks separately. This approach increases the training and deployment burden, and cannot uniformly handle modality variable inputs.

To address the above issue, we propose a switchable prompt generation (SPG) block to generate modality-aware prompts through structural switching. The modality-aware prompts drive the frozen pre-trained SOD model to address corresponding SOD tasks with few learnable parameters. It is worth noting that SPG adaptively performs structural switching based on single-modal and multi-modal inputs without manual intervention. In this way, the model can uniformly handle single-modal and multi-modal inputs end-to-end. Considering that the feature representations of different levels jointly contribute to the recognition of salient regions, we build a SPG block for each level of the extracted features to fully adapt the pre-trained model to different tasks.

\textbf{Multi-modal Prompt.}
In this case, the inputs to the model are an RGB image and a corresponding depth or thermal image. The goal of the SPG block is to exploit the complementary benefits of the two modalities to generate robust prompts. By combining them with the inherent RGB intermediate features as input to the subsequent components, the multi-modal information can be integrated to deal with complex scenes and the pre-trained model can be instructed to address the corresponding multi-modal SOD task. The upper part of Fig.~\ref{fig::SPG} illustrates the structure of the SPG for multi-modal inputs. Specifically, given the extracted features $\left\{ {f_r^i,f_a^i} \right\}_{i = 1}^4$ of two modalities in the $i$-th level, we first utilize a 3 × 3 convolutional layer to integrate the auxiliary-modal feature, which learns the properties of the unfamiliar modality (i.e., depth or thermal modality). Then, a sigmod function is employed to squeeze the feature values to the [0,1] range as a mask $\mu$, which guides the RGB feature to focus on the common information between modalities through element-wise multiplication. To further complement the information of the auxiliary modality, the original extracted auxiliary-modal feature is added to the guided RGB feature with a residual connection. Subsequently, the prompt is obtained by performing a 3 × 3 convolution on the multi-modal feature. The whole process can be formulated as:
\begin{equation}\label{Eq::mask}
	\begin{split}
		&\mu  = \sigma \left( {Con{v_{3 \times 3}}\left( {f_a^i} \right)} \right),
	\end{split}
\end{equation} 
\begin{equation}\label{Eq::multi_prompt}
	\begin{split}
		&{P^i} = Con{v_{3 \times 3}}\left( {f_r^i \odot \mu  + f_a^i} \right),
	\end{split}
\end{equation} 
where $P^i$ denotes the prompt of the $i$-th level, $\odot$ is the element-wise multiplication, and $\sigma$ is the sigmoid function. 

\textbf{Single-modal Prompt.}
In this case, the inputs to the model are two identical RGB images. The goal of the SPG block becomes to refine the inherent RGB features to generate single-modal prompts, which facilitate saliency cue mining in subsequent modules and thus drive the pre-trained model to address the RGB SOD task. In specific, since the parameters of the backbone are frozen, as the two inputs to the model become the same visible image, the two extracted features $\left\{ {f_r^i,f_r^i} \right\}_{i = 1}^4$ at each level are also the same. In this way, the structure of the SPG is switched as shown in the lower part of Fig.~\ref{fig::SPG}. The 3 × 3 convolutional layer here is used on the RGB feature to further expand the receptive field and extract more useful information, which is compressed into a mask $\mu$ by the sigmoid function. Then, the original RGB feature $f_r^i$ is guided by it through element-wise multiplication, followed by a residual connection to obtain the refined feature. The single-modal prompt $P^i$ is obtained by integrating the refined RGB feature with a 3 × 3 convolution, which can be formulated as: 
\begin{equation}\label{Eq::mask_single}
	\begin{split}
		&\mu  = \sigma \left( {Con{v_{3 \times 3}}\left( {f_r^i} \right)} \right),
	\end{split}
\end{equation} 
\begin{equation}\label{Eq::single_prompt}
	\begin{split}
		&{P^i} = Con{v_{3 \times 3}}\left( {f_r^i \odot \mu  + f_r^i} \right),
	\end{split}
\end{equation} 

\textbf{Optimization}
To optimize the prompts, we sum them with the original RGB features and send them into the multi-layer transformer encoder, the same as in~\cite{jia2022vpt}. The loss function $\cal L$ used is the same as that of the pre-trained model. During training, the data flow propagates through the entire model that contains the frozen pre-trained model and the learnable prompts, which only account for 18\% of the entire model parameters but drive the pre-trained model to achieve superior performance on both single-modal and multi-modal SOD tasks.

\begin{table*}[t]
	\centering
	\caption{Quantitative Comparison of Our Proposed UniSOD With Other 16 RGB SOD Methods on 5 Representative Datasets. The Best and Second-Best Results Are Marked With \textcolor[rgb]{ .792,  0,  0}{\textbf{Red}}, \textcolor[rgb]{ 0,  .533,  .2}{\textbf{Green}}, and \textcolor[rgb]{ 0,  .4,  .8}{\textbf{Blue}}, Respectively. The Symbols '$\uparrow$' / '$\downarrow$' Indicate That A Higher / Lower Score Is Better.}
	\resizebox{1\textwidth}{!}{
		\begin{tabular}{cc|cccccccccccccccc||c}
			\hline
			\multicolumn{2}{c|}{\multirow{2}[2]{*}{Method}} & AFNet$_{19}$ & CPD$_{19}$ & BASNet$_{19}$ & PoolNet$_{19}$ & MINet$_{20}$ & LDF$_{20}$ & GateNet$_{20}$ & F3Net$_{20}$ & PAKRN$_{21}$ & PFSNet$_{21}$ & VST$_{21}$ & EDN$_{22}$ & ICON$_{23}$ & MENet$_{23}$ & BBRF$_{23}$ & SelfReformer$_{23}$ & UniSOD \bigstrut[t]\\
			\multicolumn{2}{c|}{} &~\cite{Feng2019AFNet}       &~\cite{Wu19CPD}       &~\cite{qin2019basnet}       &~\cite{Liu2019poolnet}       &~\cite{pang20MINet}       &~\cite{Wei20LDF}       &~\cite{Zhao20GateNet}       &~\cite{Wei20F3Net}       &~\cite{Xu21KRN}       &~\cite{Ma21PFSNet}       &~\cite{liu2021vst}       &~\cite{Wu22EDN}       &~\cite{Mingchen23ICON}       &~\cite{Wang23MENet}       &~\cite{ma2023bbrf}       &~\cite{Yun23SelfReformer}       & (Ours)  \bigstrut[b]\\
			\hline
			\multicolumn{2}{c|}{Learnable params (M) \hfill$\downarrow$} & 36.0  & 47.9  & 87.1  & 68.3  & 162.4  & \textcolor[rgb]{ 0,  .533,  .2}{\textbf{25.2}} & 128.6  & \textcolor[rgb]{ 0,  .4,  .8}{\textbf{25.5}} & 45.2  & 31.2  & 32.2  & 92.4  & 42.8  & -     & 74.1  & 44.6  & \textcolor[rgb]{ .792,  0,  0}{\textbf{25.1}} \bigstrut\\
			\hline
			\multirow{4}[2]{*}{DUTS} & ${E_\xi }$ ~~~\hfill$\uparrow$   & 0.879  & 0.886  & 0.884  & 0.881  & 0.898  & 0.910  & 0.891  & 0.902  & 0.916  & 0.902  & 0.892  & \textcolor[rgb]{ 0,  .533,  .2}{\textbf{0.930}} & 0.908  & 0.921  & \textcolor[rgb]{ 0,  .4,  .8}{\textbf{0.927}} & 0.921  & \textcolor[rgb]{ .792,  0,  0}{\textbf{0.938}} \bigstrut[t]\\
			& ${S_m }$ ~~~\hfill$\uparrow$   & 0.867  & 0.869  & 0.866  & 0.879  & 0.884  & 0.892  & 0.890  & 0.888  & 0.900  & 0.892  & 0.896  & \textcolor[rgb]{ 0,  .533,  .2}{\textbf{0.917}} & 0.892  & 0.905  & 0.909  & \textcolor[rgb]{ 0,  .4,  .8}{\textbf{0.911}} & \textcolor[rgb]{ .792,  0,  0}{\textbf{0.925}} \\
			& ${F_\beta ^\omega}$ ~~~\hfill$\uparrow$   & 0.785  & 0.795  & 0.803  & 0.796  & 0.825  & 0.845  & 0.818  & 0.835  & 0.861  & 0.842  & 0.828  & \textcolor[rgb]{ 0,  .533,  .2}{\textbf{0.886}} & 0.845  & 0.870  & \textcolor[rgb]{ 0,  .533,  .2}{\textbf{0.886}} & \textcolor[rgb]{ 0,  .4,  .8}{\textbf{0.872}} & \textcolor[rgb]{ .792,  0,  0}{\textbf{0.906}} \\
			& $MAE$ \hfill$\downarrow$  & 0.046  & 0.043  & 0.048  & 0.042  & 0.037  & 0.034  & 0.038  & 0.035  & 0.033  & 0.036  & 0.037  & \textcolor[rgb]{ 0,  .533,  .2}{\textbf{0.025}} & 0.035  & 0.028  & \textcolor[rgb]{ 0,  .533,  .2}{\textbf{0.025}} & \textcolor[rgb]{ 0,  .4,  .8}{\textbf{0.027}} & \textcolor[rgb]{ .792,  0,  0}{\textbf{0.021}} \bigstrut[b]\\
			\hline
			\multirow{4}[2]{*}{ECSSD} & ${E_\xi }$ ~~~\hfill$\uparrow$   & 0.918  & 0.925  & 0.921  & 0.921  & 0.927  & 0.925  & 0.924  & 0.927  & 0.924  & 0.928  & 0.918  & \textcolor[rgb]{ 0,  .4,  .8}{\textbf{0.932}} & 0.929  & 0.925  & \textcolor[rgb]{ 0,  .533,  .2}{\textbf{0.934}} & 0.929  & \textcolor[rgb]{ .792,  0,  0}{\textbf{0.941}} \bigstrut[t]\\
			& ${S_m }$ ~~~\hfill$\uparrow$   & 0.913  & 0.918  & 0.916  & 0.917  & 0.925  & 0.924  & 0.920  & 0.924  & 0.928  & 0.930  & 0.932  & \textcolor[rgb]{ 0,  .533,  .2}{\textbf{0.941}} & 0.927  & 0.928  & \textcolor[rgb]{ 0,  .4,  .8}{\textbf{0.939}} & 0.936  & \textcolor[rgb]{ .792,  0,  0}{\textbf{0.950}} \\
			& ${F_\beta ^\omega}$ ~~~\hfill$\uparrow$   & 0.886  & 0.898  & 0.904  & 0.890  & 0.911  & 0.915  & 0.894  & 0.912  & 0.918  & 0.920  & 0.910  & \textcolor[rgb]{ 0,  .4,  .8}{\textbf{0.936}} & 0.918  & 0.920  & \textcolor[rgb]{ 0,  .533,  .2}{\textbf{0.944}} & 0.926  & \textcolor[rgb]{ .792,  0,  0}{\textbf{0.953}} \\
			& $MAE$ \hfill$\downarrow$  & 0.042  & 0.037  & 0.037  & 0.042  & 0.033  & 0.034  & 0.040  & 0.033  & 0.032  & 0.031  & 0.033  & \textcolor[rgb]{ 0,  .4,  .8}{\textbf{0.023}} & 0.032  & 0.031  & \textcolor[rgb]{ 0,  .533,  .2}{\textbf{0.022}} & 0.027  & \textcolor[rgb]{ .792,  0,  0}{\textbf{0.017}} \bigstrut[b]\\
			\hline
			\multirow{4}[2]{*}{OMRON} & ${E_\xi }$ ~~~\hfill$\uparrow$   & 0.853  & 0.866  & 0.869  & 0.857  & 0.865  & 0.874  & 0.862  & 0.870  & 0.885  & 0.875  & 0.861  & \textcolor[rgb]{ 0,  .533,  .2}{\textbf{0.898}} & 0.879  & 0.882  & \textcolor[rgb]{ 0,  .4,  .8}{\textbf{0.891}} & 0.889  & \textcolor[rgb]{ .792,  0,  0}{\textbf{0.910}} \bigstrut[t]\\
			& ${S_m }$ ~~~\hfill$\uparrow$   & 0.826  & 0.825  & 0.836  & 0.832  & 0.833  & 0.839  & 0.838  & 0.838  & 0.853  & 0.842  & 0.850  & \textcolor[rgb]{ 0,  .533,  .2}{\textbf{0.869}} & 0.849  & 0.850  & \textcolor[rgb]{ 0,  .4,  .8}{\textbf{0.861}} & \textcolor[rgb]{ 0,  .4,  .8}{\textbf{0.861}} & \textcolor[rgb]{ .792,  0,  0}{\textbf{0.876}} \\
			& ${F_\beta ^\omega}$ ~~~\hfill$\uparrow$   & 0.717  & 0.719  & 0.751  & 0.721  & 0.738  & 0.752  & 0.729  & 0.747  & 0.779  & 0.756  & 0.755  & \textcolor[rgb]{ 0,  .533,  .2}{\textbf{0.804}} & 0.770  & 0.771  & \textcolor[rgb]{ 0,  .4,  .8}{\textbf{0.803}} & 0.784  & \textcolor[rgb]{ .792,  0,  0}{\textbf{0.822}} \\
			& $MAE$ \hfill$\downarrow$  & 0.057  & 0.056  & 0.056  & 0.056  & 0.056  & 0.052  & 0.055  & 0.053  & 0.050  & 0.055  & 0.050  & \textcolor[rgb]{ 0,  .533,  .2}{\textbf{0.043}} & 0.049  & 0.045  & \textcolor[rgb]{ 0,  .4,  .8}{\textbf{0.044}} & \textcolor[rgb]{ 0,  .533,  .2}{\textbf{0.043}} & \textcolor[rgb]{ .792,  0,  0}{\textbf{0.037}} \bigstrut[b]\\
			\hline
			\multirow{4}[2]{*}{HKU-IS} & ${E_\xi }$ ~~~\hfill$\uparrow$   & 0.942  & 0.944  & 0.946  & 0.950  & 0.953  & 0.954  & 0.949  & 0.953  & 0.955  & 0.956  & 0.953  & \textcolor[rgb]{ 0,  .533,  .2}{\textbf{0.965}} & 0.956  & \textcolor[rgb]{ 0,  .4,  .8}{\textbf{0.960}} & \textcolor[rgb]{ 0,  .533,  .2}{\textbf{0.965}} & 0.959  & \textcolor[rgb]{ .792,  0,  0}{\textbf{0.969}} \bigstrut[t]\\
			& ${S_m }$ ~~~\hfill$\uparrow$   & 0.905  & 0.905  & 0.909  & 0.916  & 0.919  & 0.919  & 0.915  & 0.917  & 0.924  & 0.924  & 0.928  & \textcolor[rgb]{ 0,  .533,  .2}{\textbf{0.935}} & 0.924  & 0.927  & \textcolor[rgb]{ 0,  .4,  .8}{\textbf{0.932}} & 0.931  & \textcolor[rgb]{ .792,  0,  0}{\textbf{0.940}} \\
			& ${F_\beta ^\omega}$ ~~~\hfill$\uparrow$   & 0.869  & 0.875  & 0.889  & 0.883  & 0.897  & 0.904  & 0.880  & 0.900  & 0.909  & 0.910  & 0.897  & \textcolor[rgb]{ 0,  .4,  .8}{\textbf{0.925}} & 0.908  & 0.917  & \textcolor[rgb]{ 0,  .533,  .2}{\textbf{0.932}} & 0.915  & \textcolor[rgb]{ .792,  0,  0}{\textbf{0.939}} \\
			& $MAE$ \hfill$\downarrow$  & 0.036  & 0.034  & 0.032  & 0.032  & 0.029  & 0.028  & 0.033  & 0.028  & 0.027  & 0.026  & 0.029  & \textcolor[rgb]{ 0,  .4,  .8}{\textbf{0.022}} & 0.026  & 0.023  & \textcolor[rgb]{ 0,  .533,  .2}{\textbf{0.020}} & 0.024  & \textcolor[rgb]{ .792,  0,  0}{\textbf{0.018}} \bigstrut[b]\\
			\hline
			\multirow{4}[2]{*}{PASCAL-S} & ${E_\xi}$ ~~~\hfill$\uparrow$   & 0.851  & 0.855  & 0.852  & 0.845  & 0.857  & 0.871  & 0.857  & 0.865  & 0.863  & 0.862  & 0.843  & \textcolor[rgb]{ 0,  .4,  .8}{\textbf{0.875}} & 0.870  & 0.870  & 0.873  & \textcolor[rgb]{ 0,  .533,  .2}{\textbf{0.879}} & \textcolor[rgb]{ .792,  0,  0}{\textbf{0.887}} \bigstrut[t]\\
			& ${S_m}$ ~~~\hfill$\uparrow$   & 0.849  & 0.848  & 0.838  & 0.851  & 0.856  & 0.862  & 0.857  & 0.860  & 0.858  & 0.860  & 0.871  & \textcolor[rgb]{ 0,  .533,  .2}{\textbf{0.885}} & 0.864  & 0.871  & 0.878  & \textcolor[rgb]{ 0,  .4,  .8}{\textbf{0.881}} & \textcolor[rgb]{ .792,  0,  0}{\textbf{0.889}} \\
			& ${F_\beta ^\omega}$ ~~~\hfill$\uparrow$   & 0.798  & 0.800  & 0.798  & 0.791  & 0.815  & 0.828  & 0.803  & 0.822  & 0.823  & 0.825  & 0.822  & \textcolor[rgb]{ 0,  .4,  .8}{\textbf{0.860}} & 0.833  & 0.844  & \textcolor[rgb]{ 0,  .533,  .2}{\textbf{0.862}} & 0.854  & \textcolor[rgb]{ .792,  0,  0}{\textbf{0.875}} \\
			& $MAE$ \hfill$\downarrow$  & 0.070  & 0.070  & 0.075  & 0.071  & 0.063  & 0.059  & 0.068  & 0.062  & 0.066  & 0.063  & 0.061  & \textcolor[rgb]{ 0,  .533,  .2}{\textbf{0.048}} & 0.061  & 0.054  & \textcolor[rgb]{ 0,  .4,  .8}{\textbf{0.049}} & 0.051  & \textcolor[rgb]{ .792,  0,  0}{\textbf{0.044}} \bigstrut[b]\\
			\hline
		\end{tabular}%
	}
	\label{tab:compare_rgb}%
\end{table*}%

\begin{table*}[t]
	\centering
	\caption{Quantitative Comparison of Our Proposed UniSOD With Other 16 RGB-D SOD Methods on 6 Representative Datasets.}
	\resizebox{1\textwidth}{!}{
		\begin{tabular}{cc|cccccccccccccccc||c}
			\hline
			\multicolumn{2}{c|}{\multirow{2}[2]{*}{Method}} & HAINet$_{21}$ & SPNet$_{21}$ & RD3D$_{21}$ & DSA2F$_{21}$ & DCF$_{21}$ & VST$_{21}$ & MobileSal$_{22}$ & {SSL$_{22}$} & MoAD$_{22}$ & CIRNet$_{22}$ & DIGRNet$_{23}$ & HRTrans$_{23}$ & LSNet$_{23}$ & CAVER$_{23}$ & CATNet$_{23}$ & PICRNet$_{23}$ & UniSOD \bigstrut\\
			\multicolumn{2}{c|}{} &~\cite{Li21HAINet}       &~\cite{Zhou21SPNet}       &~\cite{Chen21RD3D}       &~\cite{sun2021dsa2f}       &~\cite{Ji21DCF}       &~\cite{liu2021vst}       &~\cite{Wu22MobileSal}       &~\cite{zhao2022self}       &~\cite{Jin22MoAD}       &~\cite{Cheng22CIRNet}    &~\cite{Cheng23DIGRNet}   &~\cite{Tang2023RTransNet}       &~\cite{zhou2023lsnet}       &~\cite{pang2023caver}       &~\cite{Sun23CATNet}       &~\cite{Cong23PICRNet}       & (Ours) \bigstrut\\
			\hline
			\multicolumn{2}{c|}{Learnable params (M) \hfill$\downarrow$} & 59.8  & 150.3  & 28.9  & 34.0  & 97.0  & 53.5  & \textcolor[rgb]{ 0,  .4,  .8}{\textbf{6.5}} & 74.2  & 103.2  & \textcolor[rgb]{ 0,  .533,  .2}{\textbf{5.0}} & 166.7  & 26.3  & \textcolor[rgb]{ .792,  0,  0}{\textbf{4.6}} & 55.8  & 263.1  & 86.0  & 25.1  \bigstrut\\
			\hline
			\multirow{2}[1]{*}{DUT-} & ${E_\xi }$ ~~~\hfill$\uparrow$   & 0.937  & 0.876  & 0.949  & 0.950  & 0.952  & 0.960  & 0.936  & 0.927  & 0.951  & 0.949  & 0.948  & \textcolor[rgb]{ 0,  .533,  .2}{\textbf{0.972}} & 0.927  & 0.955  & \textcolor[rgb]{ 0,  .4,  .8}{\textbf{0.971}} & 0.967  & \textcolor[rgb]{ .792,  0,  0}{\textbf{0.975}} \bigstrut[t]\\
			& ${S_m }$ ~~~\hfill$\uparrow$   & 0.909  & 0.803  & 0.932  & 0.921  & 0.924  & 0.943  & 0.896  & 0.889  & 0.932  & 0.927  & 0.926  & \textcolor[rgb]{ 0,  .4,  .8}{\textbf{0.951}} & 0.886  & 0.931  & \textcolor[rgb]{ .792,  0,  0}{\textbf{0.953}} & 0.943  & \textcolor[rgb]{ 0,  .533,  .2}{\textbf{0.952}} \\
			\multirow{2}[1]{*}{RGBD} & ${F_\beta ^\omega}$ ~~~\hfill$\uparrow$   & 0.887  & 0.747  & 0.913  & 0.914  & 0.913  & 0.926  & 0.869  & 0.859  & 0.908  & 0.911  & 0.902  & \textcolor[rgb]{ 0,  .533,  .2}{\textbf{0.949}} & 0.856  & 0.920  & \textcolor[rgb]{ 0,  .4,  .8}{\textbf{0.945}} & 0.935  & \textcolor[rgb]{ .792,  0,  0}{\textbf{0.952}} \\
			& $MAE$ \hfill$\downarrow$  & 0.038  & 0.085  & 0.031  & 0.030  & 0.030  & 0.025  & 0.044  & 0.046  & 0.031  & 0.031  & 0.033  & \textcolor[rgb]{ 0,  .533,  .2}{\textbf{0.018}} & 0.049  & 0.029  & \textcolor[rgb]{ 0,  .4,  .8}{\textbf{0.020}} & 0.022  & \textcolor[rgb]{ .792,  0,  0}{\textbf{0.017}} \bigstrut[b]\\
			\hline
			\multirow{4}[2]{*}{NJUD} & ${E_\xi }$ ~~~\hfill$\uparrow$   & 0.917  & \textcolor[rgb]{ 0,  .533,  .2}{\textbf{0.931}} & 0.918  & 0.923  & 0.922  & 0.913  & 0.914  & 0.881  & 0.922  & 0.909  & 0.928  & 0.929  & 0.911  & 0.922  & \textcolor[rgb]{ .792,  0,  0}{\textbf{0.932}} & \textcolor[rgb]{ 0,  .4,  .8}{\textbf{0.930}} & \textcolor[rgb]{ 0,  .4,  .8}{\textbf{0.930}} \bigstrut[t]\\
			& ${S_m }$ ~~~\hfill$\uparrow$   & 0.909  & 0.925  & 0.915  & 0.903  & 0.903  & 0.922  & 0.905  & 0.841  & 0.915  & 0.906  & \textcolor[rgb]{ .792,  0,  0}{\textbf{0.932}} & \textcolor[rgb]{ 0,  .4,  .8}{\textbf{0.926}} & 0.909  & 0.920  & \textcolor[rgb]{ .792,  0,  0}{\textbf{0.932}} & 0.924  & \textcolor[rgb]{ 0,  .533,  .2}{\textbf{0.930}} \\
			& ${F_\beta ^\omega}$ ~~~\hfill$\uparrow$   & 0.882  & \textcolor[rgb]{ 0,  .4,  .8}{\textbf{0.909}} & 0.890  & 0.889  & 0.884  & 0.892  & 0.874  & 0.786  & 0.881  & 0.881  & \textcolor[rgb]{ 0,  .4,  .8}{\textbf{0.909}} & 0.914  & 0.877  & 0.903  & \textcolor[rgb]{ 0,  .533,  .2}{\textbf{0.921}} & \textcolor[rgb]{ 0,  .4,  .8}{\textbf{0.909}} & \textcolor[rgb]{ .792,  0,  0}{\textbf{0.922}} \\
			& $MAE$ \hfill$\downarrow$  & 0.039  & 0.029  & 0.037  & 0.039  & 0.038  & 0.035  & 0.040  & 0.065  & 0.040  & 0.041  & \textcolor[rgb]{ 0,  .4,  .8}{\textbf{0.028}} & 0.029  & 0.040  & 0.032  & \textcolor[rgb]{ 0,  .533,  .2}{\textbf{0.026}} & 0.030  & \textcolor[rgb]{ .792,  0,  0}{\textbf{0.025}} \bigstrut[b]\\
			\hline
			\multirow{4}[2]{*}{NLPR} & ${E_\xi }$ ~~~\hfill$\uparrow$   & 0.951  & 0.957  & 0.957  & 0.950  & 0.956  & 0.953  & 0.950  & 0.954  & 0.955  & 0.945  & 0.955  & \textcolor[rgb]{ .792,  0,  0}{\textbf{0.968}} & 0.955  & 0.959  & \textcolor[rgb]{ 0,  .533,  .2}{\textbf{0.967}} & \textcolor[rgb]{ 0,  .4,  .8}{\textbf{0.965}} & 0.964  \bigstrut[t]\\
			& ${S_m }$ ~~~\hfill$\uparrow$   & 0.921  & 0.928  & 0.929  & 0.918  & 0.921  & 0.931  & 0.920  & 0.919  & 0.933  & 0.915  & \textcolor[rgb]{ 0,  .4,  .8}{\textbf{0.935}} & \textcolor[rgb]{ .792,  0,  0}{\textbf{0.942}} & 0.918  & 0.929  & \textcolor[rgb]{ 0,  .533,  .2}{\textbf{0.940}} & \textcolor[rgb]{ 0,  .4,  .8}{\textbf{0.935}} & 0.931  \\
			& ${F_\beta ^\omega}$ ~~~\hfill$\uparrow$   & 0.884  & 0.899  & 0.894  & 0.889  & 0.892  & 0.891  & 0.878  & 0.885  & 0.889  & 0.875  & 0.895  & \textcolor[rgb]{ .792,  0,  0}{\textbf{0.921}} & 0.881  & 0.899  & \textcolor[rgb]{ 0,  .533,  .2}{\textbf{0.916}} & \textcolor[rgb]{ 0,  .4,  .8}{\textbf{0.911}} & 0.909  \\
			& $MAE$ \hfill$\downarrow$  & 0.025  & 0.021  & 0.022  & 0.024  & 0.023  & 0.024  & 0.025  & 0.027  & 0.023  & 0.027  & 0.023  & \textcolor[rgb]{ .792,  0,  0}{\textbf{0.016}} & 0.024  & 0.022  & \textcolor[rgb]{ 0,  .533,  .2}{\textbf{0.018}} & \textcolor[rgb]{ 0,  .4,  .8}{\textbf{0.019}} & \textcolor[rgb]{ 0,  .4,  .8}{\textbf{0.019}} \bigstrut[b]\\
			\hline
			\multirow{4}[2]{*}{SSD} & ${E_\xi }$ ~~~\hfill$\uparrow$   & 0.843  & 0.910  & 0.905  & 0.904  & 0.898  & 0.907  & 0.898  & 0.833  & 0.898  & 0.894  & 0.889  & 0.910  & 0.902  & \textcolor[rgb]{ 0,  .4,  .8}{\textbf{0.915}} & \textcolor[rgb]{ 0,  .533,  .2}{\textbf{0.916}} & \textcolor[rgb]{ 0,  .4,  .8}{\textbf{0.915}} & \textcolor[rgb]{ .792,  0,  0}{\textbf{0.923}} \bigstrut[t]\\
			& ${S_m }$ ~~~\hfill$\uparrow$   & 0.769  & 0.871  & 0.863  & 0.876  & 0.852  & \textcolor[rgb]{ .792,  0,  0}{\textbf{0.889}} & 0.863  & 0.745  & 0.862  & 0.854  & 0.866  & 0.867  & 0.856  & 0.874  & \textcolor[rgb]{ 0,  .533,  .2}{\textbf{0.886}} & 0.878  & \textcolor[rgb]{ 0,  .4,  .8}{\textbf{0.880}} \\
			& ${F_\beta ^\omega}$ ~~~\hfill$\uparrow$   & 0.682  & 0.831  & 0.794  & 0.836  & 0.800  & 0.836  & 0.804  & 0.638  & 0.791  & 0.801  & 0.804  & 0.820  & 0.796  & 0.826  & \textcolor[rgb]{ 0,  .533,  .2}{\textbf{0.841}} & \textcolor[rgb]{ 0,  .4,  .8}{\textbf{0.837}} & \textcolor[rgb]{ .792,  0,  0}{\textbf{0.847}} \\
			& $MAE$ \hfill$\downarrow$  & 0.101  & 0.044  & 0.052  & 0.047  & 0.053  & 0.045  & 0.052  & 0.100  & 0.054  & 0.057  & 0.053  & 0.045  & 0.055  & \textcolor[rgb]{ 0,  .4,  .8}{\textbf{0.044}} & \textcolor[rgb]{ 0,  .533,  .2}{\textbf{0.040}} & 0.046  & \textcolor[rgb]{ .792,  0,  0}{\textbf{0.035}} \bigstrut[b]\\
			\hline
			\multirow{4}[2]{*}{SIP} & ${E_\xi }$ ~~~\hfill$\uparrow$   & 0.924  & 0.930  & 0.919  & 0.908  & 0.920  & \textcolor[rgb]{ 0,  .4,  .8}{\textbf{0.936}} & 0.914  & 0.921  & 0.917  & 0.908  & 0.918  & \textcolor[rgb]{ 0,  .533,  .2}{\textbf{0.943}} & 0.911  & 0.927  & \textcolor[rgb]{ 0,  .533,  .2}{\textbf{0.943}} & 0.916  & \textcolor[rgb]{ .792,  0,  0}{\textbf{0.945}} \bigstrut[t]\\
			& ${S_m }$ ~~~\hfill$\uparrow$   & 0.886  & 0.894  & 0.885  & 0.861  & 0.873  & 0.903  & 0.873  & 0.880  & 0.888  & 0.865  & 0.885  & \textcolor[rgb]{ 0,  .533,  .2}{\textbf{0.909}} & \textcolor[rgb]{ 0,  .533,  .2}{\textbf{0.909}} & 0.893  & \textcolor[rgb]{ .792,  0,  0}{\textbf{0.911}} & 0.865  & \textcolor[rgb]{ 0,  .4,  .8}{\textbf{0.907}} \\
			& ${F_\beta ^\omega}$ ~~~\hfill$\uparrow$   & 0.860  & 0.873  & 0.852  & 0.838  & 0.850  & 0.878  & 0.837  & 0.851  & 0.848  & 0.828  & 0.849  & \textcolor[rgb]{ 0,  .533,  .2}{\textbf{0.901}} & 0.877  & 0.874  & \textcolor[rgb]{ 0,  .4,  .8}{\textbf{0.897}} & 0.838  & \textcolor[rgb]{ .792,  0,  0}{\textbf{0.902}} \\
			& $MAE$ \hfill$\downarrow$  & 0.049  & 0.043  & 0.049  & 0.057  & 0.051  & \textcolor[rgb]{ 0,  .4,  .8}{\textbf{0.040}} & 0.054  & 0.049  & 0.053  & 0.058  & 0.053  & \textcolor[rgb]{ 0,  .533,  .2}{\textbf{0.035}} & \textcolor[rgb]{ 0,  .4,  .8}{\textbf{0.040}} & 0.043  & \textcolor[rgb]{ 0,  .533,  .2}{\textbf{0.035}} & 0.056  & \textcolor[rgb]{ .792,  0,  0}{\textbf{0.033}} \bigstrut[b]\\
			\hline
			\multirow{4}[2]{*}{STERE} & ${E_\xi }$ ~~~\hfill$\uparrow$   & 0.930  & 0.930  & 0.926  & 0.928  & \textcolor[rgb]{ 0,  .4,  .8}{\textbf{0.931}} & 0.916  & 0.916  & 0.923  & 0.921  & 0.914  & 0.927  & 0.929  & 0.913  & \textcolor[rgb]{ 0,  .4,  .8}{\textbf{0.931}} & \textcolor[rgb]{ 0,  .533,  .2}{\textbf{0.935}} & \textcolor[rgb]{ .792,  0,  0}{\textbf{0.937}} & \textcolor[rgb]{ .792,  0,  0}{\textbf{0.937}} \bigstrut[t]\\
			& ${S_m }$ ~~~\hfill$\uparrow$   & 0.909  & 0.907  & 0.911  & 0.897  & 0.905  & 0.913  & 0.903  & 0.897  & 0.891  & 0.898  & 0.916  & \textcolor[rgb]{ 0,  .533,  .2}{\textbf{0.921}} & 0.871  & 0.914  & \textcolor[rgb]{ 0,  .533,  .2}{\textbf{0.921}} & \textcolor[rgb]{ 0,  .4,  .8}{\textbf{0.920}} & \textcolor[rgb]{ .792,  0,  0}{\textbf{0.924}} \\
			& ${F_\beta ^\omega}$ ~~~\hfill$\uparrow$   & 0.877  & 0.879  & 0.877  & 0.877  & 0.880  & 0.872  & 0.865  & 0.864  & 0.836  & 0.861  & 0.877  & \textcolor[rgb]{ 0,  .4,  .8}{\textbf{0.899}} & 0.827  & 0.887  & \textcolor[rgb]{ 0,  .533,  .2}{\textbf{0.900}} & 0.898  & \textcolor[rgb]{ .792,  0,  0}{\textbf{0.908}} \\
			& $MAE$ \hfill$\downarrow$  & 0.038  & 0.037  & 0.038  & 0.038  & 0.037  & 0.038  & 0.041  & 0.042  & 0.049  & 0.042  & 0.038  & \textcolor[rgb]{ 0,  .533,  .2}{\textbf{0.030}} & 0.054  & 0.034  & \textcolor[rgb]{ 0,  .533,  .2}{\textbf{0.030}} & \textcolor[rgb]{ 0,  .4,  .8}{\textbf{0.031}} & \textcolor[rgb]{ .792,  0,  0}{\textbf{0.028}} \bigstrut[b]\\
			\hline
		\end{tabular}%
	}
	\label{tab:compare_rgbd}%
\end{table*}%

\begin{figure*}[t]
	\centering
	\includegraphics[width=1\linewidth]{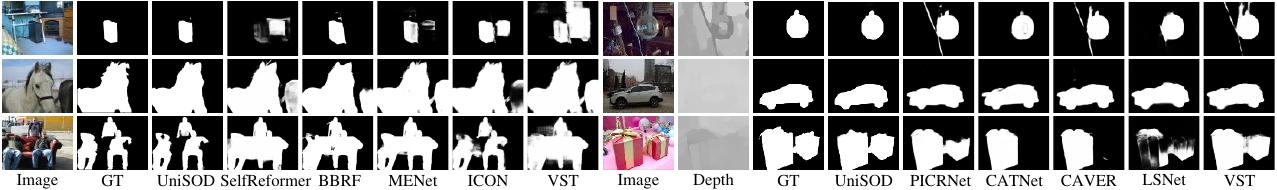}
	\caption{Qualitative comparisons of our proposed UniSOD with recent state-of-the-art RGB (left) and RGB-D (right) SOD methods in some challenging scenes. (GT: ground truth)} 
	\label{fig::RGBRGBD_show}
\end{figure*}
\section{Experiments}
\subsection{Experimental Settings}
UniSOD achieves the unification of single-modal and multi-modal SOD tasks. In this paper, we verify the effectiveness of the proposed method on three main tasks, including RGB SOD, RGB-D SOD, and RGB-T SOD.

\textbf{Datasets.} 
\textbf{(i)} For RGB SOD, we evaluate different methods on five prevalent benchmark datasets, including DUTS~\cite{wang2017Learning} (10,553 training images and 5,019 testing images), OMRON~\cite{yang2013saliency} (5,168 images), ECSSD~\cite{yan2013Hierarchical} (1,000 images), HKU-IS~\cite{li2015visual} (4,447 images), and PASCAL-S~\cite{li2014Secrets} (850 images).
\textbf{(ii)} For RGB-D SOD, all methods are assessed on six popular benchmark datasets, including DUT-RGBD~\cite{piao2019depth} (1,200 image pairs), NJUD~\cite{ju2014depth} (1,985 image pairs), NLPR~\cite{peng2014rgbd} (1,000 image pairs), SSD~\cite{zhu2017three} (80 image pairs), SIP~\cite{fan2020rethinking} (929 image pairs), and STERE~\cite{niu2012leveraging} (1,000 image pairs).
\textbf{(iii)} For RGB-T SOD, we provide the experimental results of all methods on three prevalent benchmark datasets: VT5000~\cite{tu2020rgbt} (2,500 training image pairs and 2,500 testing image pairs) (821 image pairs), VT1000~\cite{tu2019rgb} (1,000 image pairs), and VT821~\cite{tang2019rgbt}. 
During the training process, in order to reduce the storage and deployment costs, we use the training datasets commonly used for the three tasks (i.e., the training set of DUTS~\cite{wang2017Learning}, NLPR~\cite{peng2014rgbd}, NJUD~\cite{ju2014depth}, DUT-RGBD~\cite{piao2019depth}, and VT5000~\cite{tu2020rgbt}) to jointly train a unified model that handles these tasks simultaneously without specific modulations.

\begin{table}[t]
	\centering
	\caption{Quantitative Comparison of Our Proposed UniSOD With Other 6 Recent RGB-T SOD Methods on 3 Representative Datasets.}
	\resizebox{1\columnwidth}{!}{
		\begin{tabular}{cc|cccccc||c}
			\hline
			\multicolumn{2}{c|}{\multirow{2}[2]{*}{Method}} & SwinNet$_{22}$ & MCFNet$_{23}$ & HRTrans$_{23}$ & LSNet$_{23}$ & CAVER$_{23}$ & WaveNet$_{23}$ & UniSOD \bigstrut[t]\\
			\multicolumn{2}{c|}{} &~\cite{liu2022swinnet}       &~\cite{ma2023modal}       &~\cite{Tang2023RTransNet}       &~\cite{zhou2023lsnet}       &~\cite{pang2023caver}       &~\cite{zhou23WaveNet}       & (Ours) \bigstrut[b]\\
			\hline
			\multicolumn{2}{c|}{Learnable params (M) \hfill$\downarrow$} & 199.2  & 70.8  & \textcolor[rgb]{ 0,  .4,  .8}{\textbf{26.3}}  & \textcolor[rgb]{ .792,  0,  0}{\textbf{4.6}} & 55.8  & 80.7  & \textcolor[rgb]{ 0,  .533,  .2}{\textbf{25.1}}  \bigstrut\\
			\hline
			\multirow{4}[2]{*}{VT5000} & ${E_\xi}$ ~~~\hfill$\uparrow$  & \textcolor[rgb]{ 0,  .4,  .8}{\textbf{0.942}} & 0.924  & \textcolor[rgb]{ 0,  .533,  .2}{\textbf{0.945}} & 0.915  & 0.924  & 0.940  & \textcolor[rgb]{ .792,  0,  0}{\textbf{0.955}} \bigstrut[t]\\
			& ${S_m}$ ~~~\hfill$\uparrow$  & \textcolor[rgb]{ 0,  .533,  .2}{\textbf{0.912}} & 0.887  & \textcolor[rgb]{ 0,  .533,  .2}{\textbf{0.912}} & 0.877  & 0.892  & \textcolor[rgb]{ 0,  .4,  .8}{\textbf{0.911}} & \textcolor[rgb]{ .792,  0,  0}{\textbf{0.919}} \\
			& ${F_\beta ^\omega}$ ~~~\hfill$\uparrow$  & 0.846  & 0.836  & \textcolor[rgb]{ 0,  .533,  .2}{\textbf{0.870}} & 0.806  & 0.835  & \textcolor[rgb]{ 0,  .4,  .8}{\textbf{0.864}} & \textcolor[rgb]{ .792,  0,  0}{\textbf{0.891}} \\
			& $MAE$ \hfill$\downarrow$ & \textcolor[rgb]{ 0,  .4,  .8}{\textbf{0.026}} & 0.033  & \textcolor[rgb]{ 0,  .533,  .2}{\textbf{0.025}} & 0.037  & 0.032  & \textcolor[rgb]{ 0,  .4,  .8}{\textbf{0.026}} & \textcolor[rgb]{ .792,  0,  0}{\textbf{0.021}} \bigstrut[b]\\
			\hline
			\multirow{4}[2]{*}{VT1000} & ${E_\xi }$ ~~~\hfill$\uparrow$  & \textcolor[rgb]{ 0,  .4,  .8}{\textbf{0.947}}  & 0.944  & 0.945  & 0.935  & 0.945  & \textcolor[rgb]{ 0,  .533,  .2}{\textbf{0.952}} & \textcolor[rgb]{ .792,  0,  0}{\textbf{0.956}} \bigstrut[t]\\
			& ${S_m}$ ~~~\hfill$\uparrow$  & \textcolor[rgb]{ 0,  .4,  .8}{\textbf{0.938}} & 0.932  & \textcolor[rgb]{ 0,  .4,  .8}{\textbf{0.938}} & 0.925  & 0.936  & \textcolor[rgb]{ 0,  .533,  .2}{\textbf{0.945}} & \textcolor[rgb]{ .792,  0,  0}{\textbf{0.946}} \\
			& ${F_\beta ^\omega}$ ~~~\hfill$\uparrow$  & 0.894  & 0.906  & \textcolor[rgb]{ 0,  .4,  .8}{\textbf{0.913}} & 0.887  & 0.909  & \textcolor[rgb]{ 0,  .533,  .2}{\textbf{0.921}} & \textcolor[rgb]{ .792,  0,  0}{\textbf{0.931}} \\
			& $MAE$ \hfill$\downarrow$ & 0.018  & 0.019  & \textcolor[rgb]{ 0,  .4,  .8}{\textbf{0.017}} & 0.023  & \textcolor[rgb]{ 0,  .4,  .8}{\textbf{0.017}} & \textcolor[rgb]{ 0,  .533,  .2}{\textbf{0.015}} & \textcolor[rgb]{ .792,  0,  0}{\textbf{0.013}} \bigstrut[b]\\
			\hline
			\multirow{4}[2]{*}{VT821} & ${E_\xi}$ ~~~\hfill$\uparrow$  & \textcolor[rgb]{ 0,  .4,  .8}{\textbf{0.926}} & 0.918  & \textcolor[rgb]{ 0,  .533,  .2}{\textbf{0.929}} & 0.911  & 0.919  & \textcolor[rgb]{ 0,  .533,  .2}{\textbf{0.929}} & \textcolor[rgb]{ .792,  0,  0}{\textbf{0.937}} \bigstrut[t]\\
			& ${S_m}$ ~~~\hfill$\uparrow$  & 0.904  & 0.891  & \textcolor[rgb]{ 0,  .4,  .8}{\textbf{0.906}} & 0.878  & 0.891  & \textcolor[rgb]{ 0,  .533,  .2}{\textbf{0.912}} & \textcolor[rgb]{ .792,  0,  0}{\textbf{0.916}} \\
			& ${F_\beta ^\omega}$ ~~~\hfill$\uparrow$  & 0.818  & 0.835  & \textcolor[rgb]{ 0,  .4,  .8}{\textbf{0.849}} & 0.809  & 0.835  & \textcolor[rgb]{ 0,  .533,  .2}{\textbf{0.863}} & \textcolor[rgb]{ .792,  0,  0}{\textbf{0.876}} \\
			& $MAE$ \hfill$\downarrow$ & 0.030  & \textcolor[rgb]{ 0,  .4,  .8}{\textbf{0.029}} & \textcolor[rgb]{ 0,  .533,  .2}{\textbf{0.026}} & 0.033  & 0.033  & \textcolor[rgb]{ .792,  0,  0}{\textbf{0.024}} & \textcolor[rgb]{ .792,  0,  0}{\textbf{0.024}} \bigstrut[b]\\
			\hline
		\end{tabular}%
	}
	\label{tab:compare_rgbt}%
\end{table}%

\begin{figure}[t]
	\centering
	\includegraphics[width=1\linewidth]{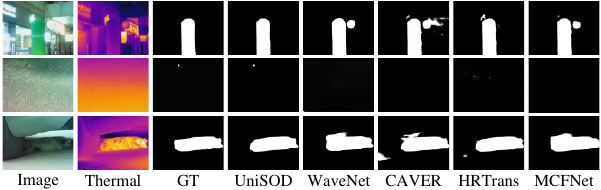}
	\caption{Qualitative comparisons of our proposed UniSOD with recent state-of-the-art RGB-T SOD methods in some challenging scenes.} 
	\label{fig::RGBT_show}
\end{figure}
\textbf{Evaluation Metrics.}
We employ four widely used evaluation metrics to assess the performance of different methods comprehensively. In specific, E-measure~\cite{fan2018enhanced} (${E_\xi}$) measures both image-level and pixel-level errors. S-measure~\cite{Cheng2021Measur} (${S_m }$) evaluates the structural similarity at the region and object level. Weighed F-measure~\cite{Margolin2014how} (${F_\beta ^\omega}$) uses weighted precision and recall to measure region-based similarity. Mean Absolute Error~\cite{Federico2012Saliency} ($MAE$) indicates the absolute error between the prediction and ground truth. We also introduce the 'precision-recall' curve to demonstrate the overall performance of the model. To assess model complexity, we also report the number of learnable parameters (params).

\textbf{Implementation Details.} 
The framework is implemented with PyTorch in a workspace with two RTX 3090 GPUs. All input images are resized into 384 × 384 for training and inferring. The backbone network is equipped with SwinB~\cite{iccv2021swin} whose parameters are trained on ImageNet~\cite{krizhevsky2017imagenet}. The AdamW~\cite{Ilya2019Regularization} algorithm with a learning rate of 1e-5 is used to optimize both the pre-trained model and the fine-tuned model. 
For the pre-trained model, we set the batch size as 4 and train for a total of 200 epochs. 
For the fine-tuned model, the frozen parameters are initialized with the pre-trained model, and the trainable parameters are random initialized. The batch size is set to 8 and a total of 300 epochs are trained for prompt learning. Apart from the pre-trained model, the proposed model works well in an end-to-end manner during the training and testing phases.

\subsection{Comparison with State-of-the-arts}
For single-modal SOD, we compare the proposed UniSOD with 16 state-of-the-art RGB SOD methods, including AFNet~\cite{Feng2019AFNet}, CPD~\cite{Wu19CPD}, BASNet~\cite{qin2019basnet}, PoolNet~\cite{Liu2019poolnet}, MINet~\cite{pang20MINet}, LDF~\cite{Wei20LDF}, GateNet~\cite{Zhao20GateNet}, F3Net~\cite{Wei20F3Net}, PAKRN~\cite{Xu21KRN}, PFSNet~\cite{Ma21PFSNet}, VST~\cite{liu2021vst}, EDN~\cite{Wu22EDN}, ICON~\cite{Mingchen23ICON}, MENet~\cite{Wang23MENet}, BBRF~\cite{ma2023bbrf}, and SelfReformer~\cite{Yun23SelfReformer}. For multi-modal SOD, we compare the proposed UniSOD with 16 state-of-the-art RGB-D SOD methods, including HAINet~\cite{Li21HAINet}, SPNet~\cite{Zhou21SPNet}, RD3D~\cite{Chen21RD3D}, DSA2F~\cite{sun2021dsa2f}, DCF~\cite{Ji21DCF}, VST~\cite{liu2021vst}, MobileSal~\cite{Wu22MobileSal}, SSL~\cite{zhao2022self}, MoAD~\cite{Jin22MoAD}, CIRNet~\cite{Cheng22CIRNet}, DIGRNet~\cite{Cheng23DIGRNet}, HRTrans~\cite{Tang2023RTransNet}, LSNet~\cite{zhou2023lsnet}, CAVER~\cite{pang2023caver}, CATNet~\cite{Sun23CATNet}, and PICRNet~\cite{Cong23PICRNet}. We also compare the proposed UniSOD with 6 recent state-of-the-art RGB-T SOD methods, including SwinNet~\cite{liu2022swinnet}, MCFNet~\cite{ma2023modal}, HRTrans~\cite{Tang2023RTransNet}, LSNet~\cite{zhou2023lsnet}, CAVER~\cite{pang2023caver}, and WaveNet~\cite{zhou23WaveNet}. 

\begin{figure}[t]
	\centering
	\includegraphics[width=1\linewidth]{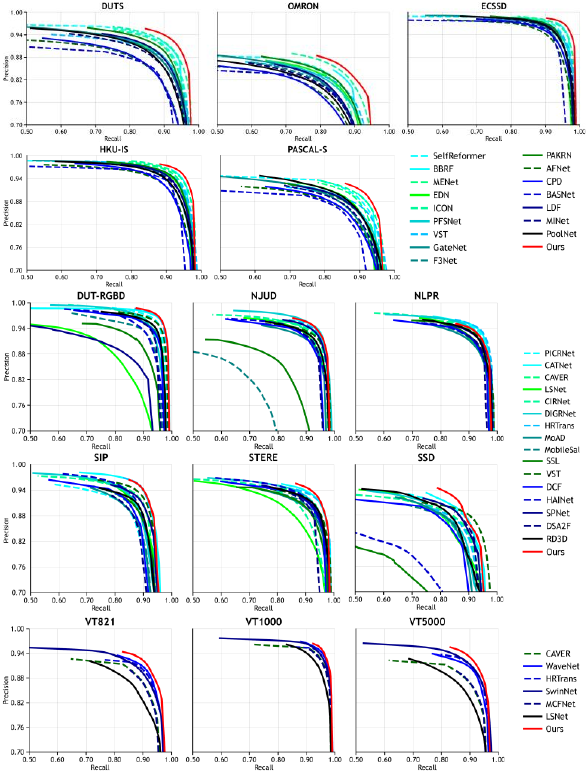}
	\caption{The Precision-Recall (PR) curves of our method and compared methods on 14 datasets.} 
	\label{fig::PR}
\end{figure}

\textbf{Quantitative Comparison.} 
Table~\ref{tab:compare_rgb}, Table~\ref{tab:compare_rgbd}, and Table~\ref{tab:compare_rgbt} show the quantitative comparison results on RGB, RGB-D, and RGB-T datasets, respectively. It can be found that our UniSOD achieves superior performance on both single-modal and multi-modal SOD tasks with a small number of learnable parameters, proving the effectiveness of our method. For example, on the RGB SOD task, compared to the sub-optimal method EDN~\cite{Wu22EDN}, our method achieves an improvement of 1.0\%, 0.7\%, 1.9\%, and 20.4\% on the four evaluation metrics (i.e., ${E_\xi }$, ${S_m }$, ${F_\beta ^\omega}$, and $MAE$) of the five datasets, respectively. In Tables~\ref{tab:compare_rgb} and~\ref{tab:compare_rgbd}, compared with VST~\cite{liu2021vst} that trains different models for RGB SOD and RGB-D SOD respectively, our method achieves better performance with only one unified model. In Tables 2 and 3, compared with the CAVER~\cite{pang2023caver} that addresses both RGB-D and RGB-T SOD, the minimum percentage gains of our UniSOD on the nine multi-modal datasets for the four evaluation metrics (i.e., ${E_\xi }$, ${S_m }$, ${F_\beta ^\omega}$, and $MAE$) are 1.5\%, 1.5\%, 2.8\%, and 34.7\%, respectively. This is mainly because the proposed SPG block can switch its structure according to the modality variable inputs to generate modal-aware prompts, which effectively drive the pre-trained model to address the corresponding SOD tasks. Furthermore, Fig.~\ref{fig::PR} shows the PR curves of our method and the compared methods on all 14 datasets. It shows that the curve of our method is further outward, indicating that our method has higher confidence and accuracy for salient regions.

\begin{table*}[t]
	\centering
	\caption{Ablation Studies of Our Proposed UniSOD on Five Single-Modal and Multi-Modal Datasets. '$w/o$' Means To Remove the Component. The Best Results Are Marked With \textbf{Bold}.}
	\resizebox{1\textwidth}{!}{
		\begin{tabular}{c|c|cccc|cccc|cccc|cccc|cccc}
			\hline
			\multirow{2}[4]{*}{Model} & Learnable & \multicolumn{4}{c|}{DUTS}     & \multicolumn{4}{c|}{ECSSD}    & \multicolumn{4}{c|}{STERE}    & \multicolumn{4}{c|}{NJUD}     & \multicolumn{4}{c}{VT5000} \bigstrut\\
			\cline{3-22}          & params (M) \hfill$\downarrow$ & ${E_\xi }$ \hfill$\uparrow$    & ${S_m }$ \hfill$\uparrow$    & ${F_\beta ^\omega}$ \hfill$\uparrow$     & $MAE$ \hfill$\downarrow$   & ${E_\xi }$ \hfill$\uparrow$    & ${S_m }$ \hfill$\uparrow$    & ${F_\beta ^\omega}$ \hfill$\uparrow$     & $MAE$ \hfill$\downarrow$   & ${E_\xi }$ \hfill$\uparrow$    & ${S_m }$ \hfill$\uparrow$    & ${F_\beta ^\omega}$ \hfill$\uparrow$     & $MAE$ \hfill$\downarrow$   & ${E_\xi }$ \hfill$\uparrow$    & ${S_m }$ \hfill$\uparrow$    & ${F_\beta ^\omega}$ \hfill$\uparrow$     & $MAE$ \hfill$\downarrow$   & ${E_\xi }$ \hfill$\uparrow$    & ${S_m }$ \hfill$\uparrow$    & ${F_\beta ^\omega}$ \hfill$\uparrow$     & $MAE$ \hfill$\downarrow$ \bigstrut\\
			\hline
			Baseline SOD model & 144.6  & 0.937  & 0.923  & 0.904  & \textbf{0.021} & 0.939  & 0.948  & 0.951  & 0.019  & 0.905  & 0.903  & 0.879  & 0.038  & 0.909  & 0.913  & 0.900  & 0.034  & 0.944  & 0.902  & 0.868  & 0.026  \bigstrut[t]\\
			UniSOD & \textbf{25.1}  & \textbf{0.938} & \textbf{0.925} & \textbf{0.906} & \textbf{0.021} & \textbf{0.941} & \textbf{0.950} & \textbf{0.953} & \textbf{0.017} & \textbf{0.937} & \textbf{0.924} & \textbf{0.908} & \textbf{0.028} & \textbf{0.930} & \textbf{0.930} & \textbf{0.922} & \textbf{0.025} & \textbf{0.955} & 0.919  & \textbf{0.891}  & \textbf{0.021} \bigstrut[b]\\
			\hline
			ICON$_{23}$~\cite{Mingchen23ICON} & 92.4  & 0.930  & 0.917  & 0.886  & 0.025  & 0.932  & 0.941  & 0.936  & 0.023  & 0.901  & 0.902  & 0.873  & 0.041  & 0.904  & 0.910  & 0.888  & 0.038  & 0.941  & 0.898  & 0.852  & 0.028  \bigstrut[t]\\
			UniSOD\_ICON & \textbf{25.1}  & 0.933  & 0.918  & 0.892  & 0.023  & 0.936  & 0.945  & 0.942  & 0.021  & 0.933  & 0.922  & 0.899  & 0.030  & 0.925  & 0.927  & 0.911  & 0.029  & 0.943  & 0.913  & 0.868  & 0.025  \bigstrut[b]\\
			\hline
			$w/o$ SPG & \textbf{25.1}  & 0.934  & 0.922  & 0.901  & 0.022  & 0.938  & 0.946  & 0.948  & 0.020  & 0.921  & 0.919  & 0.893  & 0.032  & 0.919  & 0.921  & 0.896  & 0.034  & 0.950  & 0.916  & 0.881  & 0.023  \bigstrut[t]\\
			Full fine-tuning & 169.7  & 0.936  & 0.924  & \textbf{0.906} & \textbf{0.021} & 0.937  & 0.948  & 0.951  & 0.018  & 0.928  & 0.922  & 0.904  & 0.029  & 0.926  & 0.928  & 0.915  & 0.026  & 0.953  & \textbf{0.921} & 0.890  & 0.023  \bigstrut[b]\\
			\hline
		\end{tabular}%
	}
	\label{tab:ablation}%
\end{table*}%

\textbf{Qualitative Comparison.} 
Figs.~\ref{fig::RGBRGBD_show} and~\ref{fig::RGBT_show} illustrate the visual comparison between our UniSOD and some advanced methods. The results show that our method is able to deal with different challenging scenes, such as complex background, scale variation, modality failure, etc. For example, for the RGB SOD sample in the first row of Fig.~\ref{fig::RGBRGBD_show}, our method can accurately locate the salient region with little noise in a complex background. For the RGB-T SOD example in the second row of Fig.~\ref{fig::RGBT_show}, our method is able to recognize and segment the salient object even when the object is tiny and the auxiliary modality fails. These results demonstrate the robustness of our UniSOD on different data, which is attributed to the strong information modeling capability of the proposed SPG block for both single-modal and multi-modal inputs.

\subsection{Ablation Study}
In this section, we perform ablation studies to illustrate the effectiveness of the components in our method. The results on single-modal (i.e., DUTS and ECSSD) and multi-modal (i.e., STERE, NJUD, and VT5000) datasets are shown in Table~\ref{tab:ablation}, in which the second line (i.e., UniSOD) shows the performance of our full model.

\textbf{Effectiveness of the baseline SOD model.}
To verify the effectiveness of the proposed pre-trained baseline SOD model, we directly evaluate its performance on both single-modal and multi-modal datasets using only visible images as input, denoted as 'Baseline SOD model' in Table~\ref{tab:ablation}. Compared with the advanced RGB SOD methods in Table~\ref{tab:compare_rgb}, it can be found that our pre-trained model has superior performance with a simple structure and acceptable parameters, which indicates that it has great potential for exploration. Compared to the multi-modal methods in Tables~\ref{tab:compare_rgbd} and~\ref{tab:compare_rgbt}, the advantages of our pre-trained model are not obvious, which confirms that a specific single-modal SOD model is difficult to handle multi-modal SOD tasks well at the same time. In addition, the results in the first and second rows of Table~\ref{tab:ablation} indicate that the superior performance of our UniSOD is inseparable from the rich saliency prior knowledge provided by the pre-trained baseline SOD model.

We also replace the proposed baseline SOD model with an existing advanced single-modal SOD model (i.e., ICON~\cite{Mingchen23ICON}), which also uses SwinB~\cite{iccv2021swin} as the backbone. The results are demonstrated in the second and third rows of Table~\ref{tab:ablation}, in which 'UniSOD\_ICON' indicates that the ICON was used as a frozen pre-trained model driven by the modality-aware prompts to address both single-modal and multi-modal SOD tasks. By comparison, UniSOD\_ICON exhibits a weaker performance than UniSOD, which indicates the effectiveness of the proposed baseline SOD model. In addition, the average improvement of UniSOD over the proposed baseline SOD model is 21.5\%, while the average improvement of UniSOD\_ICON over ICON is 19.6\% on the $MAE$ metric across the five datasets. This is mainly due to the simple and effective structure of the proposed baseline SOD model, which has great potential to fully interact with modality-aware prompts and the pre-trained model.

\begin{figure}[t]
	\centering
	\includegraphics[width=1\linewidth]{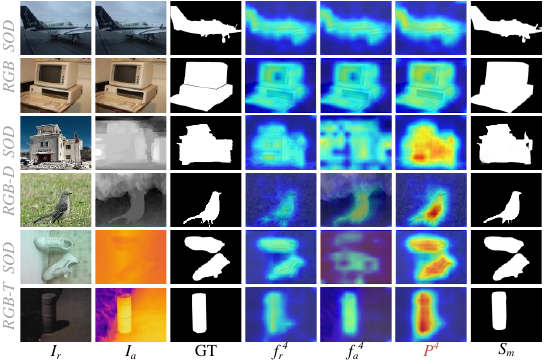}
	\caption{Feature visualization for the SPG block of the highest layer. $I_r$ and $I_a$ are input images, $f_r^4$ and $f_a^4$ are input features for SPG, $P^4$ is the output prompt of SPG, $S_m$ is the prediction, and GT is ground truth.} 
	\label{fig::visual}
\end{figure}
\begin{figure}[t]
	\centering
	\includegraphics[width=1\linewidth]{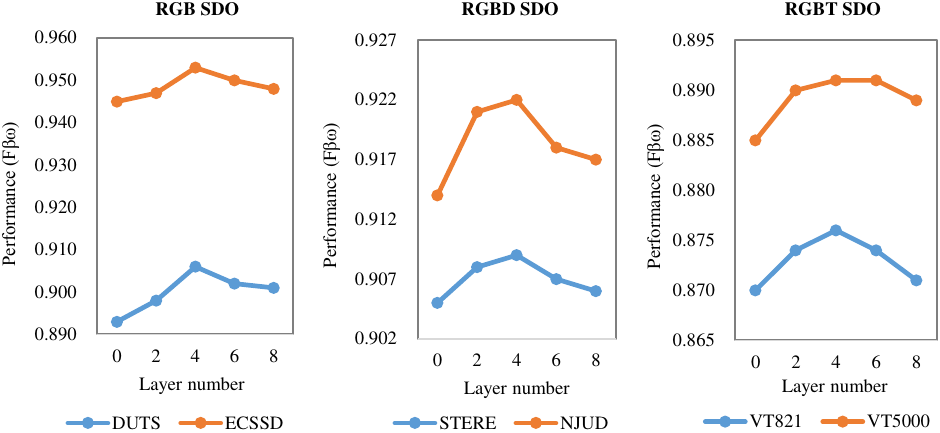}
	\caption{Effect of the number of transformer encoder layers on the proposed UniSOD.} 
	\label{fig::layer}
\end{figure}
\textbf{Effectiveness of SPG.}
To verify the effectiveness of SPG and guarantee fairness, we replace the SPG block with two consecutive convolutional blocks to ensure that the learnable parameters are invariant, which means that the model is not able to switch structures in response to modality variable inputs. The results in the fifth row (i.e., '$w/o$ SPG') of Table~\ref{tab:ablation} verify the positive effect of the SPG block. Compared with UniSOD, the performance of the four metrics (i.e., ${E_\xi }$, ${S_m }$, ${F_\beta ^\omega}$, and $MAE$) on the five datasets drops by an average of 0.8\%, 0.5\%, 1.4\%, and 16.4\% respectively. 
It is worth noting that its performance is also inferior to the full fine-tuning variant (i.e. the sixth row in Table~\ref{tab:ablation}), which is inferior to our UniSOD. This shows that the SPG block can generate more effective prompts to drive the pre-trained model to achieve superior performance. In addition, UniSOD and UniSOD\_ICON in Table~\ref{tab:ablation} both have significant improvements compared with their respective pre-trained models (i.e., baseline SOD model and ICON), which illustrates the positive effect of SPG.

Fig.~\ref{fig::visual} further vividly illustrates the visualization of the input features ($f_r^4$ and $f_a^4$) and output features ($P^4$) of the SPG block. For single-modal input, the SPG block enhances the representation of input features through refinement, such as object boundaries and highlights, which prompts the pre-trained model to distinguish foreground regions more accurately, as shown in the first and second rows of Fig.~\ref{fig::visual}. For multi-modal inputs, the SPG block exploits the complementary benefits of the two modalities to drive the pre-trained model to deal with challenging scenes, such as auxiliary modality failure (i.e., third and fifth rows) and RGB modality being disturbed (i.e., forth and sixth rows).

\textbf{Effectiveness of prompt learning.}
In order to observe the effectiveness of prompt learning, we fine-tune the entire model that contains the pre-trained model and the SPG block, with results shown in the fourth row (i.e., 'Full fine-tuning') of Table~\ref{tab:ablation}. It can be found that its performance is inferior to our UniSOD despite having more learnable parameters. This is mainly because plenty of parameters require more training data to learn them well. On the other hand, full fine-tuning breaks the well learned prior knowledge of the pre-trained model.

\textbf{Effectiveness of Transformer Layers.}
Since the blended input with the appended modality-aware prompts mainly interacts in the transformer encoder, we verify the effect of the transformer encoder layers on our UniSOD. Fig.~\ref{fig::layer} illustrates the ${F_\beta ^\omega}$ performance variation of UniSOD on both single-modal and multi-modal representative datasets. As the number of layers increases, the modality-aware prompts can interact with the inherent input more adequately to adapt the pre-trained model better to the corresponding task. However, when the number of layers exceeds 4, the performance shows a decreasing trend. This is mainly because more layers result in a larger number of parameters, but the limited training data makes it difficult to fit. Therefore, we adopt a 4-layer transformer encoder setting here.

\begin{table}[t]
	\centering
	\caption{Ablation Studies of Our Proposed UniSOD on Joint training and Task-Specific Training. UniSOD\_X Represents Training the Modality-Aware Prompt Using the Specific Training Dataset of the X SOD Task. The Best Results Are Marked With \textbf{Bold}.}
	\resizebox{1\columnwidth}{!}{
		\begin{tabular}{c|c|cccc|cccc}
			\hline
			\multirow{2}[4]{*}{RGB SOD}   & Learnable & \multicolumn{4}{c|}{DUTS}     & \multicolumn{4}{c}{ECSSD} \bigstrut\\
			\cline{3-10}     & params (M) & ${E_\xi }$ \hfill$\uparrow$    & ${S_m }$ \hfill$\uparrow$    & ${F_\beta ^\omega}$ \hfill$\uparrow$     & $MAE$ \hfill$\downarrow$   & ${E_\xi }$ \hfill$\uparrow$    & ${S_m }$ \hfill$\uparrow$    & ${F_\beta ^\omega}$ \hfill$\uparrow$     & $MAE$ \hfill$\downarrow$ \bigstrut\\
			\hline
			UniSOD & 25.1  & \textbf{0.938} & \textbf{0.925} & \textbf{0.906} & \textbf{0.021} & \textbf{0.941} & \textbf{0.950} & \textbf{0.953} & \textbf{0.017} \bigstrut[t]\\
			UniSOD\_RGB & 25.1 (total 75.3)  & 0.937  & 0.924  & 0.905  & \textbf{0.021} & 0.940  & 0.949  & 0.952  & 0.018  \bigstrut[b]\\
			\hline
			\multirow{2}[4]{*}{RGB-D SOD} & Learnable & \multicolumn{4}{c|}{STERE}    & \multicolumn{4}{c}{NJUD} \bigstrut\\
			\cline{3-10}    & params (M) & ${E_\xi}$ \hfill$\uparrow$    & ${S_m }$ \hfill$\uparrow$    & ${F_\beta ^\omega}$ \hfill$\uparrow$     & $MAE$ \hfill$\downarrow$   & ${E_\xi }$ \hfill$\uparrow$    & ${S_m }$ \hfill$\uparrow$    & ${F_\beta ^\omega}$ \hfill$\uparrow$     & $MAE$ \hfill$\downarrow$ \bigstrut\\
			\hline
			UniSOD & 25.1  & 0.937  & \textbf{0.924} & 0.908  & 0.028  & 0.930  & 0.930  & 0.922  & \textbf{0.025} \bigstrut[t]\\
			UniSOD\_RGB-D & 25.1 (total 75.3)  & \textbf{0.938} & \textbf{0.924} & \textbf{0.909} & \textbf{0.027} & \textbf{0.940} & \textbf{0.932} & \textbf{0.925} & \textbf{0.025} \bigstrut[b]\\
			\hline
			\multirow{2}[4]{*}{RGB-T SOD} & Learnable & \multicolumn{4}{c|}{VT5000}   & \multicolumn{4}{c}{VT821} \bigstrut\\
			\cline{3-10}    & params (M) & ${E_\xi }$ \hfill$\uparrow$    & ${S_m }$ \hfill$\uparrow$    & ${F_\beta ^\omega}$ \hfill$\uparrow$     & $MAE$ \hfill$\downarrow$   & ${E_\xi }$ \hfill$\uparrow$    & ${S_m }$ \hfill$\uparrow$    & ${F_\beta ^\omega}$ \hfill$\uparrow$     & $MAE$ \hfill$\downarrow$ \bigstrut\\
			\hline
			UniSOD & 25.1  & \textbf{0.955} & \textbf{0.919} & \textbf{0.891} & \textbf{0.021} & \textbf{0.937} & \textbf{0.916} & \textbf{0.876} & \textbf{0.024} \bigstrut[t]\\
			UniSOD\_RGB-T & 25.1 (total 75.3)  & 0.953  & \textbf{0.919} & 0.889  & \textbf{0.021} & 0.932 & 0.911 & 0.868 & 0.026 \bigstrut[b]\\
			\hline
		\end{tabular}%
	}
	\label{tab:GenVsSpe}%
\end{table}%
\subsection{Experiments on Specific and Joint Training}
UniSOD is trained by combining the training datasets of three tasks, including RGB SOD, RGB-D SOD, and RGB-T SOD. For a comprehensive comparison, following~\cite{zhou2023texture,sun2021dsa2f,tu2022weakly}, we use the training set of DUTS~\cite{wang2017Learning} to train a specific RGB SOD model, a collection of 2985 samples from NLPR~\cite{peng2014rgbd}, NJUD~\cite{ju2014depth}, and DUT-RGBD~\cite{piao2019depth} to train a specific RGB-D model, and the training set of VT5000~\cite{tu2020rgbt} to train a specific RGB-T model. The results on five representative datasets are shown in Table~\ref{tab:GenVsSpe}, where UniSOD denotes our full model with joint training and UniSOD\_X denotes the model trained by our method on each specific X SOD training set. It is worth noting that the total number of parameters required for the specific training is the sum of parameters for each specific model.
By comparing the performance of the three specific models with the performance of other advanced models in Tables~\ref{tab:compare_rgb},~\ref{tab:compare_rgbd}, and~\ref{tab:compare_rgbt}, it can be found that the specific models of our method still achieve a superior performance, which further validates the effectiveness of our method. In addition, our UniSOD achieves comparable performance to each specific model on its respective task with fewer parameters, which validates the effectiveness of SPG in structural switching of single-modal and multi-modal inputs, as well as the fairness of joint training for our method.

\begin{table*}[t]
	\centering
	\caption{Quantitative Comparison on Five Representative Datasets Through Jointly Training Four Advanced Multi-Modal Methods and Two Advanced Single-Modal Methods. The Best Results Are Marked with \textbf{Bold}.}
	\resizebox{1\textwidth}{!}{
		\begin{tabular}{c|c|c|cccc|cccc|cccc|cccc|cccc}
			\hline
			\multicolumn{2}{c|}{\multirow{2}[4]{*}{Model}} & Learnable & \multicolumn{4}{c|}{DUTS}     & \multicolumn{4}{c|}{ECSSD}    & \multicolumn{4}{c|}{STERE}    & \multicolumn{4}{c|}{NJUD}     & \multicolumn{4}{c}{VT5000} \bigstrut\\
			\cline{4-23}    \multicolumn{2}{c|}{} & params (M) \hfill$\downarrow$ & ${E_\xi }$ \hfill$\uparrow$    & ${S_m }$ \hfill$\uparrow$    & ${F_\beta ^\omega}$ \hfill$\uparrow$     & $MAE$ \hfill$\downarrow$   & ${E_\xi }$ \hfill$\uparrow$    & ${S_m }$ \hfill$\uparrow$    & ${F_\beta ^\omega}$ \hfill$\uparrow$     & $MAE$ \hfill$\downarrow$   & ${E_\xi }$ \hfill$\uparrow$    & ${S_m }$ \hfill$\uparrow$    & ${F_\beta ^\omega}$ \hfill$\uparrow$     & $MAE$ \hfill$\downarrow$   & ${E_\xi }$ \hfill$\uparrow$    & ${S_m }$ \hfill$\uparrow$    & ${F_\beta ^\omega}$ \hfill$\uparrow$     & $MAE$ \hfill$\downarrow$   & ${E_\xi }$ \hfill$\uparrow$    & ${S_m }$ \hfill$\uparrow$    & ${F_\beta ^\omega}$ \hfill$\uparrow$     & $MAE$ \hfill$\downarrow$ \bigstrut\\
			\hline
			Unified-modal & UniSOD & 25.1  & \textbf{0.938} & \textbf{0.925} & \textbf{0.906} & \textbf{0.021} & \textbf{0.941} & \textbf{0.950} & \textbf{0.953} & \textbf{0.017} & \textbf{0.937} & \textbf{0.924} & \textbf{0.908} & \textbf{0.028} & \textbf{0.930} & \textbf{0.930} & \textbf{0.922} & \textbf{0.025} & \textbf{0.955} & \textbf{0.919} & \textbf{0.891} & \textbf{0.021} \bigstrut\\
			\hline
			\multirow{4}[2]{*}{Multi-modal} & CAVER$_{23}$~\cite{pang2023caver} & 55.8  & 0.893  & 0.877  & 0.822  & 0.041  & 0.918  & 0.920  & 0.906  & 0.035  & 0.934  & 0.912  & 0.889  & 0.035  & 0.927  & 0.922  & 0.913  & 0.030  & 0.928  & 0.901  & 0.848  & 0.032  \bigstrut[t]\\
			& LSNet$_{23}$~\cite{zhou2023lsnet} & \textbf{4.6} & 0.869  & 0.850  & 0.770  & 0.054  & 0.917  & 0.910  & 0.888  & 0.042  & 0.916  & 0.886  & 0.832  & 0.052  & 0.916  & 0.914  & 0.883  & 0.038  & 0.917  & 0.882  & 0.809  & 0.035  \\
			& HRTrans$_{23}$~\cite{Tang2023RTransNet} & 26.3  & 0.921  & 0.912  & 0.878  & 0.027  & 0.932  & 0.942  & 0.937  & 0.023  & 0.931  & 0.921  & 0.901  & 0.031  & 0.928  & 0.927  & 0.917  & 0.028  & 0.950  & 0.921  & 0.885  & 0.023  \\
			& SwinNet$_{22}$~\cite{liu2022swinnet} & 199.2 & 0.912  & 0.908  & 0.861  & 0.029  & 0.932  & 0.943  & 0.936  & 0.024  & 0.926  & 0.919  & 0.878  & 0.036  & 0.920  & 0.928  & 0.897  & 0.032  & 0.948  & \textbf{0.919} & 0.865  & 0.023  \bigstrut[b]\\
			\hline
			\multirow{2}[2]{*}{Single-modal} & EDN$_{22}$~\cite{Wu22EDN} & 42.8  & 0.932  & 0.920  & 0.884  & 0.024  & 0.935  & 0.946  & 0.938  & 0.021  & 0.914  & 0.898  & 0.861  & 0.042  & 0.909  & 0.906  & 0.881  & 0.041  & 0.874  & 0.821  & 0.716  & 0.045  \bigstrut[t]\\
			& BBRF$_{23}$~\cite{ma2023bbrf} & 74.1  & 0.929  & 0.912  & 0.889  & 0.025  & 0.935  & 0.942  & 0.942  & 0.021  & 0.916  & 0.903  & 0.866  & 0.041  & 0.914  & 0.905  & 0.874  & 0.040  & 0.898  & 0.864  & 0.763  & 0.043  \bigstrut[b]\\
			\hline
		\end{tabular}%
	}
	\label{tab:jointCom}%
\end{table*}%
In Table~\ref{tab:jointCom}, we further verify the impact of joint training on the performance of six existing state-of-the-art methods, including CAVER~\cite{pang2023caver}, LSNet~\cite{zhou2023lsnet}, HRTrans~\cite{Tang2023RTransNet}, SwinNet~\cite{liu2022swinnet}, EDN~\cite{Wu22EDN}, and BBRF~\cite{ma2023bbrf}. Specifically, we train these models using the code published by the authors and their default settings. By comparing the results of these methods in Table~\ref{tab:jointCom} with their corresponding original results in Tables~\ref{tab:compare_rgb},~\ref{tab:compare_rgbd}, and~\ref{tab:compare_rgbt}, it can be found that these methods improve the overall performance on their specific tasks. For example, the average improvement of CAVER on the four evaluation metrics (i.e., ${E_\xi }$, ${S_m }$, ${F_\beta ^\omega}$, and $MAE$) across the three datasets (i.e., STERE, NJUD, and VT5000) is 0.4\%, 0.3\%, 1.0\%, and 1.3\%. On the other hand, these methods struggle to achieve superior performance on the datasets beyond their specific tasks. For example, although EDN achieves outstanding performance on the single-modal task, it has no obvious performance advantage on the multi-modal tasks. These results show that joint training can indeed bring performance improvements to the models, especially on their specific tasks, but it cannot help the model directly achieve uniformly excellent performance on all tasks.
In addition, it can be found that our UniSOD still achieves better performance across all the five datasets, which also illustrates its effectiveness.

\begin{figure*}[t]
	\centering
	\includegraphics[width=1\linewidth]{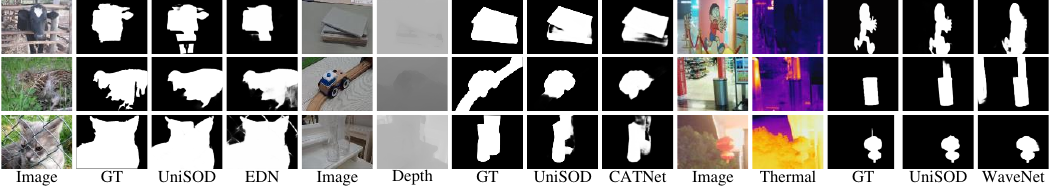}
	\caption{Visualization of some typical failure cases of our UniSOD and several advanced methods~\cite{Wu22EDN,Sun23CATNet,zhou23WaveNet} on RGB, RGB-D, and RGB-T test datasets.} 
	\label{fig::failure}
\end{figure*}
\subsection{Failure Cases and Future Works}
Although our method achieves superior performance on both single-modal and multi-modal SOD tasks, it still fails in some extremely challenging scenarios, as shown in Fig~\ref{fig::failure}. For the RGB SOD samples on the left, the object of interest is difficult to be accurately localized due to occlusion. The RGB-D and RGB-T SOD samples on the right demonstrate the difficulty of capturing salient objects in the case of poor auxiliary modality. This is mainly due to the fact that these challenges are significantly difficult, and our UniSOD, which is a unified and general framework for multiple tasks, lacks a challenge-specific design. Likewise, existing advanced methods also have difficulty in addressing these challenges.
To address this issue, we plan to add corresponding designs to the baseline SOD model in the future work, which can provide prior knowledge for downstream tasks to deal with these challenges. In addition, by optimizing the design of the SPG block for the characteristics of different tasks, prompts for specific challenges can be generated for the pre-trained model. 

\section{Conclusion}
In this paper, we are the first to address both single-modal and multi-modal SOD tasks in a unified framework. To handle the difficulty of input modality variations, we propose the UniSOD to learn modality-aware prompts through adaptive prompt learning with few learnable parameters, which can drive pre-trained models to address the corresponding tasks. To this end, two components are proposed. The baseline SOD model provides rich prior knowledge for both single-modal and multi-modal downstream SOD tasks with a simple and effective structure. The SPG block is cleverly designed to perform the structural switching based on single-modal and multi-modal inputs. Experimental results on 14 benchmark datasets show that our UniSOD achieves superior performance for RGB, RGB-D, and RGB-T SOD, which demonstrates its great effectiveness and potential.

\bibliographystyle{IEEEtran}
\bibliography{mybibfile}

\end{document}